\documentclass[journal]{IEEEtran}

\IEEEoverridecommandlockouts                              

\usepackage{amsfonts,amsmath} 

\usepackage{tikz}
\usetikzlibrary{arrows,calc,decorations.pathmorphing,decorations.markings,backgrounds,decorations,decorations.pathmorphing,positioning,fit,automata,shapes,patterns,decorations.text}

\usepackage{tabulary,multirow}
\newcolumntype{K}[1]{>{\centering\arraybackslash}p{#1}}

\usepackage{outlines}
\let\labelindent\relax
\usepackage{enumitem}
\setenumerate[1]{label=\Roman*.}
\setenumerate[2]{label=\Alph*.}
\setenumerate[3]{label=\roman*.}
\setenumerate[4]{label=\alph*.}

\usepackage{graphicx} 
\graphicspath{{./figures/}}
\usepackage{color}  
\usepackage{xspace}
\usepackage{subfig}
\usepackage{url}

\newcommand{\toolname}{Sim-ATAV\xspace}
\newcommand{\GUR}{Global UR\xspace}
\newcommand{\CAUR}{CA+UR\xspace}
\newcommand{\CASA}{CA+SA\xspace}

\newcommand{\matlab}{MATLAB\textsuperscript{\circledR}\xspace}

\newcommand{\reals}{\ensuremath{\mathbb{R}}}
\renewcommand{\int}{\ensuremath{\mathbb{Z}}}

\newcommand{\model}{\ensuremath{\Sigma}}
\newcommand{\modeld}{\ensuremath{M_d}}
\newcommand{\modela}{\ensuremath{M_1}}
\newcommand{\modelb}{\ensuremath{M_2}}
\newcommand{\modelc}{\ensuremath{M_3}}
\newcommand{\phid}{\ensuremath{\phi_d}}
\newcommand{\X}{\ensuremath{\mathcal{X}}}

\newcommand{\Y}{\ensuremath{\mathcal{Y}}}
\newcommand{\Yb}{\ensuremath{{\bf Y}}}
\newcommand{\collestimated}{\ensuremath{FC}}

\newcommand{\y}{\ensuremath{{\bf y}}}
\newcommand{\U}{\ensuremath{\mathcal{U}}}
\newcommand{\Ub}{\ensuremath{ {\bf U}}}
\renewcommand{\u}{\ensuremath{{\bf u}}}
\renewcommand{\P}{\ensuremath{\mathcal{P}}}
\newcommand{\p}{\ensuremath{{\bf p}}}
\newcommand{\Pc}{\ensuremath{\mathcal{W}}}
\newcommand{\pc}{\ensuremath{{\bf w}}}
\newcommand{\Pd}{\ensuremath{\mathcal{V}}}
\newcommand{\pd}{\ensuremath{{\bf v}}}

\renewcommand{\sim}{\ensuremath{sim}}

\newcommand{\inpTr}{ { \ensuremath{ \bf u} } }
\newcommand{\simTr}{ { \ensuremath{ \pmb{\sigma}} } }

\newcommand{\squeezedet}{SqueezeDet\xspace}
\newcommand{\staliro}{S-TaLiRo\xspace}

\newcommand{\Ic}{\mathcal{I}}
\newcommand{\Oc}{\mathcal{O}}
\newcommand{\dle}{[\![} 
\newcommand{\dri}{]\!]} 
\newcommand \genMet { d }
 
\newcommand \robsemD[1] {\dle #1 \dri_{\genMet}} 
\newcommand{\tss}{\simTr}
 
\newcommand \copreals {\reals_{\geq 0}}

\def\checkmark{\tikz\fill[scale=0.4](0,.35) -- (.25,0) -- (1,.7) -- (.25,.15) -- cycle;}

\newcommand{\systemreq}{\ensuremath{R1}} 
\newcommand{\sensorIreq}{\ensuremath{R2}} 
\newcommand{\sensorIIreq}{\ensuremath{R3}} 
\newcommand{\sensetosystemreq}{\ensuremath{R4}} 
\newcommand{\perfreq}{\ensuremath{R5}} 
\newcommand{\edge}{\ensuremath{B^{\downarrow}}} 

\newtheorem{example}{Example}
\newtheorem{remark}{Remark}

\newtheorem{definition}{Definition}

\usepackage{ifthen}
\newboolean{revVer}
\setboolean{revVer}{false}
\newboolean{inclAuthors}
\setboolean{inclAuthors}{false}

\title{\LARGE \bf Requirements-driven Test Generation for Autonomous Vehicles with Machine Learning Components*}

\author{Cumhur Erkan Tuncali$^{1}$, Georgios Fainekos$^{1}$, Danil Prokhorov$^{2}$, Hisahiro Ito$^{2}$, and James Kapinski$^{2}$
\thanks{*This work was partially funded by NSF awards CNS 1446730, 1350420}
\thanks{$^{1}$School of Computing, Informatics \& Decision Systems Engineering, Arizona State University, USA {\tt\small etuncali, fainekos@asu.edu}}%
\thanks{$^{2}$Toyota Technical Center, Ann Arbor MI, USA {\tt\small danil.prokhorov@toyota.com}}%
}

\begin{document}

\ifthenelse {\boolean{revVer}}
{
\newcommand{\hlRev}[1] { \textcolor{blue}{#1} }
}
{
\newcommand{\hlRev}[1] { {#1} }
}

\maketitle
\thispagestyle{empty}
\pagestyle{empty}

\begin{abstract}

Autonomous vehicles are complex systems that are challenging to test and debug.
A requirements-driven approach to the development process can decrease the resources required to design and test these systems, while simultaneously increasing the reliability.
We present a testing framework that uses signal temporal logic (STL), which is a precise and unambiguous requirements language.
Our framework evaluates test cases against the STL formulae and additionally uses the requirements to automatically discover test cases that fail to satisfy the requirements.
One of the key features of our tool is the support for machine learning (ML) components in the system design, such as deep neural networks. 
The framework allows evaluation of the control algorithms, including the ML components, and it also includes models of CCD camera, lidar, and radar sensors, as well as the vehicle environment.
We use multiple methods to generate test cases, including covering arrays, which is an efficient method to search discrete variable spaces.
The resulting test cases can be used to debug the controller design by identifying controller behaviors that do not satisfy requirements. The test cases can also enhance the testing phase of development by identifying critical corner cases that correspond to the limits of the system's allowed behaviors.
We present STL requirements for an autonomous vehicle system, which capture both component-level and system-level behaviors.
Additionally, we present three driving scenarios and demonstrate how our requirements-driven testing framework can be used to identify critical system behaviors, which can be used to support the development process.

\end{abstract}

\section{Introduction} \label{sec:intro}

Autonomous driving systems are in a stage of rapid research and development spanning a broad range of maturity from simulations to on-road testing and deployment. 
They are expected to have a significant impact on the vehicle market and the broader economy and society in the future.

Testing of highly automated and autonomous driving systems is also an area of active research.  
Both governmental and non-governmental organizations are grappling with the unique requirements of these new, highly complex systems, as they have to operate safely and reliably in diverse driving environments.  
Government and industry sponsored partnerships have produced a number of guiding documents and 
clarifications, such as NHTSA \cite{NHTSA2016}, SAE \cite{SAE2015}, CAMP \cite{Christensen2015}, NCAP \cite{NCAP2018}, PEGASUS \cite{Pegasus2018}.
The research community has also been contributing to the development of
methodologies for testing automated driving systems. 

Stellet et al. \cite{Stellet2015TestingOA} surveyed existing approaches to testing such as
simulation-only, X-in-the-loop and augmented reality approaches, as
well as test criteria and metrics (see also \cite{Zofka2018}). 
Koopman and Wagner identified challenges of
testing and proposed potential solutions, such as fault injection, as a
way to perform more efficient edge case testing \cite{Koopman2016}. 
The publications \cite{Burgard2016} and \cite{Wachenfeld2016} provide in-depth discussions on the challenges of safety validation for autonomous vehicles, arguing that virtual testing should be the main target for both methodological and economic reasons.

However, no universally agreed upon testing or verification methods have yet arisen for autonomous driving systems. 
One reason is that the current autonomous systems architectures usually include some Machine Learning (ML) components, such as Deep Neural Networks (DNNs), which are notoriously difficult to test and verify. 
For instance, Automated Driving System (ADS) designs often use ML components such as DNNs to classify objects within CCD images and to determine their positions relative to the vehicle, a process known as \emph{object detection and classification} \cite{geiger2012we,wu2017squeezedet}. 
Other designs use Neural Networks (NNs) to perform \emph{end-to-end} control of the vehicle, meaning that the NN takes in the image data and outputs actuator commands, without explicitly performing an intermediate object detection step \cite{pomerleau1989alvinn,chen2015deepdriving,Chi2017}. Still other approaches use end-to-end learning to do intermediate decisions like risk assessment~\cite{StricklandFBA2018icra}. 

ML system components are problematic from an analysis perspective, as it is difficult or impossible to characterize all of the behaviors of these components under all circumstances. One reason is that the complexity of these systems can be very high in terms of the number of parameters. For example, AlexNet \cite{krizhevsky2012imagenet}, a pre-trained DNN that is used for classification of CCD images, has 60 million parameters. Another reason for the difficulty in characterizing behaviors of ML components is that the parameters are learned based on training data. 
In other words, characterizing ML behaviors is, in some ways, as difficult as the task of characterizing the training data. 
Again using the AlexNet example, the number of training images used was 1.2 million. 
While the main strength of DNNs is their ability to generalize from training data, the major challenge for analysis is that we do not understand well how they generalize to all possible cases.
Therefore, there has been significant interest  on verification and testing for ML components. 
For example, adversarial testing approaches seek to identify perturbations in image data that result in misclassifications \cite{TianEtAl2018icse,PapernotEtAl2017asiaccs,WickerHK2018tacas}. 
However, most of the existing work on testing and verification of systems with ML components focuses only on the ML components themselves, without consideration of the closed-loop behavior of the system. 

The closed-loop nature of a typical autonomous driving system can be described as follows. 
A \emph{perception} system processes data gathered from various sensing devices, such as cameras, lidar, and radar. 
The output of the perception system is an estimation of the principal (\emph{ego}) vehicle's position with respect to external obstacles (e.g., other vehicles, called \emph{agent} vehicles, and pedestrians). 
A \emph{path planning} algorithm uses the output of the perception system to produce a short-term plan for how the ego vehicle should behave. 
A \emph{tracking controller} then takes the output of the path planner and produces actuation outputs, such as accelerator, braking, and steering commands. The actuation commands affect the vehicle's interaction with the environment. 
The iterative process of sensing, processing, and actuating is what we refer to as closed-loop behavior.

\hlRev{It is important to note that by design, closed-loop systems have error tolerance mechanisms.
Hence, adversarial attacks that work on individual ML components may not the same effect on the closed loop system.}
Since for ADS applications the ultimate goal is to evaluate the closed-loop system performance, any testing methods used to evaluate such systems should support this goal.

We present a framework for Simulation-based Adversarial Testing of Autonomous Vehicles (\toolname),  which can be used to check closed-loop properties of ADS that include ML components.
In particular, our work focuses on methods to determine perturbations in the configuration of a test scenario, meaning that we seek to find scenarios that lead to unexpected behaviors, such as misclassifications and ultimately vehicle collisions. 
The framework that we present allows this type of testing in a virtual environment.
By utilizing advanced 3D models and image rendering tools, such as the ones used in game engines, the gap between testing in a virtual environment and the real world can be minimized.
We describe a testing methodology, based on a test case generation method, called covering arrays \cite{hartman2005software}, and requirement falsification methods \cite{AbbasFSIG13tecs} to automatically identify problematic test scenarios. 
The resulting framework can be used to increase the reliability of autonomous driving systems.

An earlier version of this work appeared in \cite{Tuncali2018}.
The contributions of that work can be summarized as follows. 
In \cite{Tuncali2018}, we provided a new algorithm to perform falsification of formal requirements for an autonomous vehicle in a closed-loop with the perception system, which includes an efficient means of searching over discrete and continuous parameter spaces. The method represents a new way to do adversarial testing in scenario \emph{configuration space}, as opposed to the usual method, which considers adversaries in \emph{image space}. Additionally, we demonstrated a new way to characterize problems with perception systems in \emph{configuration space}. Lastly, we extended the software testing theory of covering arrays to closed-loop Cyber-Physical System (CPS) applications that have embedded ML algorithms.

The present paper provides the following contributions that are in addition to those from \cite{Tuncali2018}:

\begin{itemize}
\item We add models of lidar and radar sensors and include sensor fusion algorithms, and we demonstrate how the requirements-based testing framework we propose can be used to automate the search for specific types of fault cases involving sensor interactions. 

\item We provide requirements for both component-level and system-level behaviors, and 
we show how to automate the identification of behaviors where component-level failures lead to system-level failures.
An example of the kind of analysis this allows is  
automatically finding cases where a sensor failure leads to a collision case.

\item We include a model of agent \emph{visibility} to various sensors and include this notion in the requirements that we consider. This provides a way to reason about how the system should behave, based on whether agents are or are not visible, including the ability to reason about the temporal aspects of agent visibility. For example, we can use this feature to test the requirement that within 1 second after an agent becomes visible to the lidar sensor, the perception system should correctly classify the agent.
This allows us to automate the search for behaviors related to temporal aspects of sensor behaviors in the context of a realistic driving scenario.

\item We demonstrate the ability to falsify properties by adversarially searching over agent trajectories.
This permits the use of our requirements-driven search-based approach over a broad class of agent behaviors, which allows us to automatically identify corner cases that are difficult to find using traditional simulation-based techniques.

\item \hlRev{We have released a publicly available toolbox \toolname as an add-on to the \staliro falsification toolbox for Matlab$^\circledR$ \cite{AnnapureddyLFS11tacas}: \url{https://sites.google.com/a/asu.edu/s-taliro/s-taliro/sim-atav}}

\end{itemize}


\section{Related work} \label{sec:related}

Testing and evaluation methods for Autonomous%
\footnote{We utilize the more general term ``autonomous" as opposed to a more restricted ``automated" since our methods could potentially apply to all levels of autonomy.}
Vehicles (AVs) could be categorized into three major classes: (1) model based, (2) data-driven, and (3) scenario based.
Scenario-based approaches utilize accident reports and driving conditions that are easily identifiable as challenging, producing specific test scenarios to be executed either in the real world or in a simulation environment.
For example, Euro NCAP \cite{NCAP2018} and DOT \cite{NajmEtAl2013dot} provide such scenarios.
Data-driven approaches, on the other hand, typically utilize driving data \cite{ZhaoGJ17itsc} to generate probabilistic models of human drivers.
Such models are then used for risk assessment and rare event sampling for AV algorithms under specific driving scenarios \cite{ZhaoEtAl2017its}.

The aforementioned testing methods are important and necessary before AV deployment, but they cannot help with design exploration and automated fault detection at early development stages.
Such problems are addressed by model-based verification \cite{LoosPN11fm,AlthoffEtAl2010ivs}, model based test generation \cite{TuncaliPF16itsc,TuncaliF19itsc,AlthoffL2018iv,TuncaliEtAl2017case,OKelly2017,KimKDS17esl,KimJSSY16emsoft}, or a combination thereof \cite{FanQM2018ieeedt,KellyAM2016}. 
It is important to also highlight that these methods typically ignore or use simple models to abstract away proximity sensors and, especially, the vision systems.
However, ignoring sensors or using simplified sensing models may be a dangerously simplifying assumption since it ignores the complex interactions between the dynamics of the vehicle and the sensors.
For example, the effective sensing range of a sensor platform mounted on the roof of a vehicle is affected when the vehicle makes hard turns.

In addition, vision-based perception systems have become an integral component of the sensor platform of AVs, and in many cases, they constitute the only perception system.
Currently, the winning algorithmic technology for image processing systems is utilizing DNNs.
For instance, by 2011, the DNN architecture proposed in \cite{ciresan11} was already capable of classifying pre-segmented images of traffic signs with better accuracy than humans ($99.46\%$ vs $99.22\%$).
Since then, there has been substantial progress with DNNs performing both segmentation and classification  \cite{john14,angelova15icra}.
Yet, in spite of the multiple impressive results using DNNs, it is still also easy to devise methods that can produce (so-called adversarial) images that will fool them
\cite{TianEtAl2018icse,PapernotEtAl2017asiaccs}. 

The latter (negative) result raises two important questions: 
(1) can we still generate adversarial inputs for DNNs when we manipulate the physical properties and trajectories of the objects in the environment of the AV, and 
(2) how does the DNN accuracy affect the system level properties of an AV, that is, its functional safety?
Exhaustive verification methods for DNNs in-the-loop are still in their infancy \cite{DuttaEtAl2018adhs}, and they cannot handle AVs with DNN components in the loop.
To address the two questions above, several model-based test generation methods have been proposed  \cite{DreossiDS2017nfm,Dreossi2018,Abbas17Cyphy,Tuncali2018}.
The procedure described in \cite{DreossiDS2017nfm,Dreossi2018} analyzes the performance of the perception system using static images to identify candidate counterexamples, which are then checked using simulations of the closed-loop system to determine whether the AV exhibits unsafe behaviors.
On the other hand, \cite{Abbas17Cyphy,Tuncali2018} develop methods that directly search for unsafe behaviors of the closed-loop system by defining a cost function on the closed-loop behaviors.
The differences between \cite{Abbas17Cyphy} and \cite{Tuncali2018} are primarily on the search methods, the simulation environments, and the AVs, with \cite{Tuncali2018} providing a more efficient method for combinatorial search.

In this extended version of \cite{Tuncali2018}, we take the system-level adversarial test generation methods for AV one step further.
We demonstrate that our framework \cite{Tuncali2018} can be extended for test generation for AV with multi-sensor systems as opposed to vision-only perception systems.
Moreover, we demonstrate the importance and effectiveness of test generation methods guided by system-level requirements as well as perception-level requirements.

Using our framework, we can formalize and test against requirements on the sensor performance in the context of a driving scenario.
For example, the lidar's point cloud density drops significantly with the distance to the target object, for example, a pedestrian.
Similar to this aspect of lidar behavior, the pixel count of a CCD camera would also decrease dramatically with the distance if it were to be used for pedestrian detection, since the area of an observed object decreases as the square of the distance to the object.  
This may complicate testing for long-range observation conditions. Our framework supports testing these aspects of sensor performance.

\section{Preliminaries} \label{sec:preliminaries}

This section presents the setting used to describe the testing procedures performed with our framework. The purpose of our framework is to provide a mechanism to test, evaluate, and improve on an autonomous driving system design. To do this, we use a simulation environment that incorporates models of a vehicle (called the \emph{ego} vehicle), a perception system, which is used to estimate the state of the vehicle with respect to other objects in its environment, a controller, which makes decisions about how the vehicle will behave, and the environment in which the ego vehicle is deployed. The environment model contains representations of a wide variety of objects that can interact with the ego vehicle, including roads, buildings, pedestrians, and other vehicles (called \emph{agent} vehicles). The behaviors of the system are determined by the evolution of the model states over time, which we compute using a \emph{simulator}. 

\hlRev{
In the following, $\reals$ represents the set of real numbers, while $\int$ the set of integers.
In addition, $\overline{\reals} = \reals \cup \{ \pm \infty\}$ and $\reals_{\geq 0}$ is the set of positive reals.
Formally, we assume that a test scenario $\model$ is captured by a simulation function $\sim : \X_0 \times \Ub \times \P \times \reals_{\geq 0} \rightarrow \Yb$ that maps a vector of initial conditions $x_0 \in \X_0$, a vector of parameters $p\in \P$, a total simulation time $T \in \reals_{\geq 0}$ and a time stamped input signal $\u \in \Ub$ to a time stamped output signal $\y \in \Yb$. 
}

\hlRev{
Here, $\X_0 \subseteq \reals^{n^x_1} \times \int^{n^x_2}$ is the set of initial conditions for the whole scenario, i.e., for the ego vehicle(s) as well as any other stateful object in the environment.
The variable $n^x_1$ captures the number of the continuous-valued state variables in the system (i.e., the order of the differential and/or difference equations), and $n^x_2$ captures the number of discrete-valued (and, primarily, finite-valued) state variables in the system.
In other words, we assume that the models we consider are hybrid dynamical systems \cite{Alur15book}.
Similarly, $\P \subseteq \reals^{n^p_1} \times \int^{n^p_2}$ is a set of $n^p_1$ continuous-valued parameters, such as ambient temperature, light intensity, or color, and $n^p_2$ discrete-valued (categorical) parameters, such as vehicle model or sign type. 
}

\hlRev{
The set of potential input values is denoted by $\U \subseteq \reals^{m_{in}}$, where $m_{in}$ is the number of time varying input signals to the test scenario. 
The set of possible input signals is $\Ub = (\U \times \reals_{\geq 0})^{N_{in}+1}$, where $N_{in}$ is the number of samples for the input signal.
In other words, an {\it input signal} (also referred to as {\it input trace}) $ \inpTr \in \Ub$ is a function $\inpTr : \{0, 1, \ldots, N_{in} \} \rightarrow \U \times \reals_{\geq 0}$ which maps each sample $i$ to an input value $u_i \in \U$ and a time stamp $t_i \in \reals_{\geq 0}$.
Alternatively, we can view $\inpTr$ as a finite sequence of input signal values and their corresponding times:
\[\inpTr=(u_0,t_0)(u_1,t_1)\cdots (u_{N_{in}},t_{N_{in}}).\] 
Here, we make two assumptions : (i) $t_{N_{in}} $ is no greater than the simulation time $T$ (i.e., $t_{N_{in}} \leq T$), and (ii) the timestamps are monotonically increasing: $\forall i, j \in \{0, 1, \ldots, N_{in} \}$, if $i<j$, then $t_i < t_j$.
Finally, since the simulator may need to produce output values at some time $t$ between two timestamps (i.e., $t \in (t_i, t_{i+1})$), we will assume that the simulator decides what interpolation function it will use. 
}

\hlRev{
Given initial conditions $x_0 \in \X_0$, parameter values $p \in \P$, input signals $\inpTr \in \Ub$, and the total simulation time $T$, the simulator returns an {\it output trajectory} (also referred to as {\it output trace}) $\y = \sim(x_0,\inpTr,p,T)$.
The output trace is a function $\y : \{0, 1, \ldots, N_{out} \} \rightarrow \Y \times \reals_{\geq 0}$ that maps each sample $i$ to an output value $y_i \in \Y \subseteq \reals^{m_{out}}$ and a time stamp $t_i \in \reals_{\geq 0}$.
Here, $m_{out}$ is the number of observable output variables.
We denote the set of possible output traces by $\Yb = (\Y \times \reals_{\geq 0})^{N_{out}}$.
The output trace timestamps should satisfy (i) $t_{N_{out}} = T$, and (ii) the monotonicity property.
}

\hlRev{
For notational convenience, we will make an additional assumption that the simulator also returns an updated input trace $\inpTr$ where $N = N_{in}= N_{out}$ and the timestamps of $\y$ and $\inpTr$ match.
We refer to the triple $\simTr = (\y,\inpTr,p)$ as a {\it simulation trace}, which can also be viewed as a function $\simTr : \{0, 1, \ldots, N_{out} \} \rightarrow \Y  \times \U \times \P \times \reals_{\geq 0}$, or as a sequence (recall that $p$ is constant): 
 \[ \simTr = (y_0,u_0,p,t_0)(y_1,u_1,p,t_1) \cdots (y_N,u_N,p,t_N). \]
We denote the set of all simulation traces $\simTr$ of $\model$ by $\mathcal{L}(\model)$.
}

\begin{remark}
\hlRev{
		In this paper, the function $\sim$ is assumed to be deterministic; however, the results we present are also applicable to stochastic systems (i.e., when the $\sim$ function is stochastic). 
		See \cite{AbbasHFU14cyber} for a discussion.
}
\end{remark}

\begin{figure}[t]
	\begin{center}
		\includegraphics[width=0.9\linewidth]{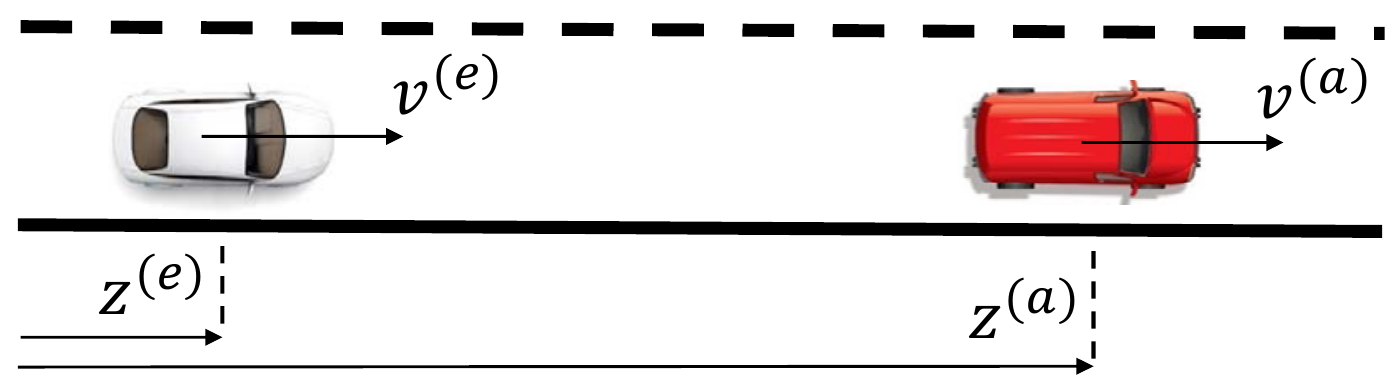} 
	\end{center}
	\caption{The simple test scenario of Example \ref{exmp:simple:2car}.}
	\label{fig:simple:exmp2car}
\end{figure}

\begin{figure}[t]
	\begin{center}
		\includegraphics[width=1\linewidth]{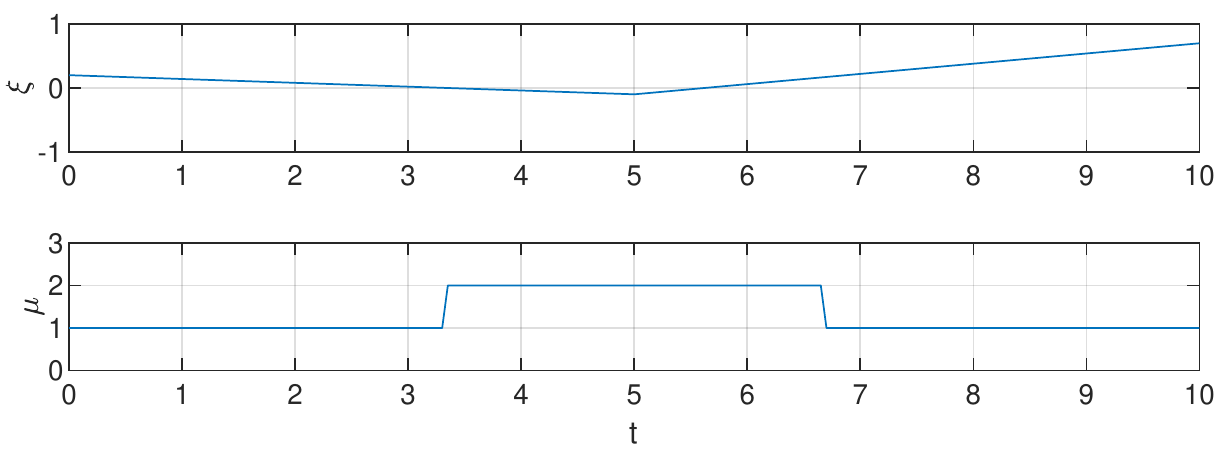} 
	\end{center}
	\vspace*{-10pt}
	\caption{Input trace $u$ for the test scenario of Example \ref{exmp:simple:2car}.}
	\label{fig:simple:exmp2car:inp}
\end{figure}

\begin{figure}[t]
	\begin{center}
		\includegraphics[width=1\linewidth]{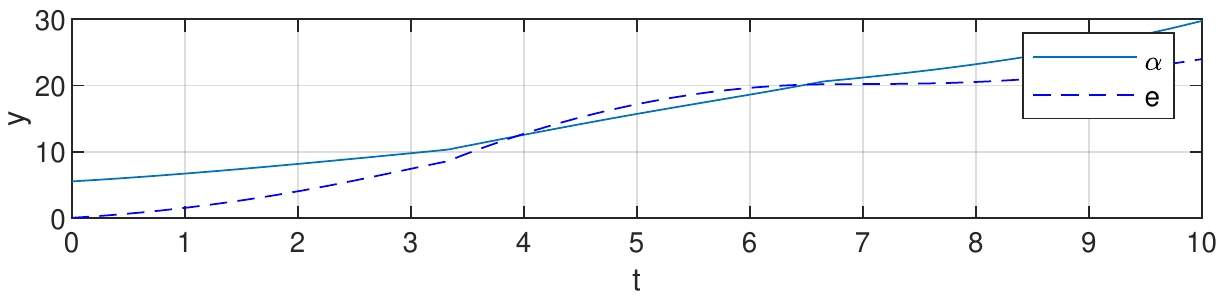} 
	\end{center}
	\vspace*{-10pt}
	\caption{Output trace $y$ for the test scenario of Example \ref{exmp:simple:2car}.}
	\label{fig:simple:exmp2car:out}
\end{figure}

\begin{example}
	\label{exmp:simple:2car}
\hlRev{
We will present a simple illustrative example to clarify the notation. 
Let's assume a test scenario as in Fig. \ref{fig:simple:exmp2car}.	
For $i \in \{a,e\}$, we will denote by $z^{(i)}$ the longitudinal position of the vehicle $i$ and by $v^{(i)}$ the velocity of the vehicle $i$.
The ego vehicle ($e$) implements an adaptive cruise control (ACC) algorithm, which we treat as a black box: $ \dot \eta^{(e)} = f_e (\eta^{(e)},\eta^{(a)})$, where $\eta^{(i)} = [z^{(i)} \; v^{(i)}]^T$.
In this simple model of an ego car, the ego car senses its environment by measuring the state $\eta^{(a)}$ of the adversarial agent. 
The adversarial agent ($a$) has simple integrator dynamics $\dot z^{(a)} = \mu v^{(a)}$ and $\dot v^{(a)} = \xi$ (with the additional constraint of no negative velocity, i.e., $v(t)\geq 0$), where $\xi$ and $\mu$ are the time varying inputs that we search over, i.e., $u = [\xi \; \mu]^T$.
In particular, $\mu(t) \in \{1,2\}$ models the normal versus the sport driving mode in the powertrain (selected by the driver), while $\xi(t) = [-1,1]$ is the acceleration (and braking) input also provided by the driver of the adversarial vehicle.
Assuming $N_{in}=201$, then $\Ub = ( [-1,1] \times \{1,2\} )^{201}$ -- see Fig. \ref{fig:simple:exmp2car:inp} for an example input. 
In this test scenario, the state space is $x = [ z^{(e)} \; v^{(e)} \; z^{(a)} \; v^{(a)}]^T$ 
and, thus $\X_0$ is the set of initial positions $z^{(i)}$ and velocities $v^{(i)}$ of the two vehicles.
The set of parameters is empty since this test scenario does not have any constant parameters.
The output trace is defined to be the positions of the two vehicles over time, i.e., $y = [ z^{(e)} \; z^{(a)} ]^T$.
Fig. \ref{fig:simple:exmp2car:out} presents the vehicle positions over time for the inputs in Fig. \ref{fig:simple:exmp2car:inp}.
It can be observed that the ego vehicle does not utilize a safe ACC since it collides with the adversarial vehicle at about time 4.
}
\end{example}

\subsection{Signal Temporal Logic}
\label{sec:mtl:intro}

Signal Temporal Logic (STL) was introduced as a syntactic extension to Metric Temporal Logic (MTL) to reason about real-time properties of signals (simulation traces) (for an overview see \cite{BartocciEtAl2018survey}).
STL formulae are built over predicates on the variables of a signal using Boolean and temporal operators.  
The temporal operators include {\it eventually} $(\Diamond_\Ic)$, {\it always} $(\Box_\Ic)$ and {\it until} $(U_\Ic)$,
where $\Ic$ is a time interval that encodes timing constraints.  
The boolean operators include {\it conjunction} $\wedge$, {\it disjunction} $\vee$, {\it negation} $\neg$, and {\it implication} $\implies$.

In this work, we interpret STL formulas over the observable simulation traces.
STL specifications can describe the usual properties of interest in system design such as (bounded time) {\bf reachability}, for example, {\it eventually, between time 1.2 and 5 (not including), $y$ should drop below $-10$}: $\Diamond_{[1.2,5)}\ ( y \leq -10)$, and {\bf safety}, for example, {\it after time 2 time units, $y$ should always be greater than $10$}: $\Box_{[2,+\infty)} (y \geq 10)$.  
\hlRev{
An important class of expressible requirements in STL are {\bf reactive requirements}, such as $\Box ( (y\leq -10) \rightarrow \Diamond_{[0,2]} (y\geq 10))$, which states that {\it whenever $y$ drops below -10, then within 2 time units $y$ should rise above $10$}.
}

Informally speaking, we allow predicate expressions to capture arbitrary constraints over the output variables, inputs, and parameters of the system.
More formally, we assume that predicates $\pi$ are expressions built using the grammar $\pi ::= f(y,u,p)\geq c \; | \; \neg \pi_1 \; | \; (\pi) \; | \; \pi_1 \vee \pi_2 \; | \; \pi_1 \wedge \pi_2$, where $f$ is a function and $c$ is a constant in $\reals$.
In other words, each predicate $\pi$ represents a subset in the space $\Y \times \U \times \P$.
In the following, we represent the set that corresponds to the predicate $\pi$ using the notation $\Oc(\pi)$. 
For example, if $\pi = (y^{(1)} \leq -10) \vee (y^{(1)} +  y^{(2)} \geq 10)$ where $y^{(i)}$ is the $i$-th component of the vector $y$, then $\Oc(\pi ) = (\infty,-10] \times \reals \cup \{y \in \reals^2 \; | \; y^{(1)} +  y^{(2)} \geq 10 \}$. 

\begin{definition}[STL Syntax] 
	Assume $\Pi$ is the set of predicates and $\Ic$ is any non-empty connected interval of $\copreals$.
	The set of all well-formed STL formulas is inductively defined as
	$ \varphi \; ::= \; \top \; | \; \pi \; | \; \neg \phi \; | \; \phi_1 \vee \phi_2 \; | \; \bigcirc\phi\; | \;\phi_1 U_\Ic \phi_2$,
	where $\pi$ is a predicate, $\top$ is \emph {true}, $\bigcirc$ is Next, and $U_\Ic$ is the Until operator.
\end{definition}

In this work, we will be using discrete time semantics of STL since we would like to be able to reason about the timing of samples and define events as falling or raising Boolean values using the next time operator ($\bigcirc$).
For example, the formula $\pi \wedge \bigcirc \neg \pi$ expresses an event (e.g., high to low).
For STL formulas $\psi$, $\phi$, we define $\psi\wedge\phi\equiv\neg(\neg\psi\vee\neg\phi)$, $\bot\equiv\neg\top$ (False), $\psi\rightarrow\phi\equiv\neg\psi\vee\phi$ ($\psi$ Implies $\phi$), $\Diamond_I\psi\equiv\top U_I\psi$ (Eventually $\psi$), $\Box_I\psi\equiv\neg\Diamond_I\neg\psi$ (Always~$\psi$), and $\psi R_I\phi\equiv\neg(\neg\psi U_I\neg\phi)$ ($\psi$ Releases $\phi$), using syntactic manipulation.


In our previous work \cite{FainekosP09tcs}, we proposed robust semantics for STL formulas.  
Robust semantics (or robustness metrics) provide a real-valued measure of satisfaction of a formula by a trace. 
In contrast, Boolean semantics just provide a {\it true} or {\it false} valuation.  
In more detail, given a trace $\simTr$ of the system, its robustness w.r.t. a temporal property $\varphi$, denoted $\robsemD{\varphi}(\simTr)$, yields a positive value if $\simTr$ satisfies $\varphi$ and a negative value otherwise.
Moreover, if the trace $\simTr$ satisfies the specification $\phi$, then the robust semantics evaluate to the radius of a neighborhood such that any other trace that remains within that neighborhood also satisfies the same specification.
The same holds for traces that do not satisfy~$\phi$. 
\hlRev{
	In order to define neighborhoods for requirement satisfaction, we need to utilize metrics over the space of outputs, inputs and parameters.
}

\begin{definition}[Metric] 
\hlRev{
	A metric on a set $S$ is a positive function $\genMet : S \times S \rightarrow \reals_{\geq 0}$ such that 
	\begin{enumerate}
		\item $\forall s, s' \in S$, $\genMet(s,s') = 0 \Leftrightarrow s =s'$ \label{cond:metr1}
		\item $\forall s, s' \in S$, $\genMet(s, s') = \genMet(s', s)$
		\item $\forall s, s', s'' \in S$, $\genMet(s, s'')\leq \genMet(s, s') + \genMet(s', s'')$
	\end{enumerate}
}
\end{definition}

\hlRev{
	In this paper, all the experimental results were derived using the Euclidean metric, i.e., $\genMet(s,s') = \| s-s' \|$; however, any other metric could be used.
	Using a metric, we can define a distance function that will capture how {\it robustly} a point belongs to a set. 
	That is, the further away a point is from the boundary of the set, the more robust is its membership in that set.
}

\begin{definition}[Signed Distance \cite{BoydV_book04} $\S 8$] 
\hlRev{
	Let $s \in S$ be a point, $A \subseteq S$ be a set and $\genMet$ be a metric on $S$. Then, we define the Signed Distance from $s$ to $A$ to be 
		\[ \mathbf{Dist}_\genMet(s,A) := \left\{ \begin{array}{ll}
		-\inf\{d(s,s')\;|\;s' \in A \} & \mbox{ if } s \not \in A \\
		\inf\{d(s,s')\;|\;s' \not \in A \} & \mbox{ if } s \in A \\
		\end{array} \right. \]
}
\end{definition}

\hlRev{
That is, the signed distance is positive if the point is in the set and negative otherwise.
}

\begin{definition} [STL Robust Semantics] 
	Given a metric $\genMet$, trace $\tss$, and $\Oc : \Pi \rightarrow 2^{\Y \times \U \times \P}$, the robust semantics of any formula $\phi $ w.r.t $\tss$ at time instance $i\in N$ is defined as:
	\begin{align*} 
	\vspace*{-0.2in}
	\allowdisplaybreaks
	\dle \top\dri_{\genMet} (\tss,i) := &  +\infty  
	\displaybreak[2] \\
	\hlRev{ \dle \pi \dri_{\genMet} (\tss,i) :=} & \hlRev{\mathbf{Dist}_\genMet([y_i \, u_i \, p_i]^T, \Oc(\pi)) }
	\displaybreak[2] \\
	\dle \neg \phi \dri_{\genMet}  (\tss,i)  :=  & - \dle \phi \dri_{\genMet}  (\tss,i) 
	\displaybreak[2] \\
	\dle \phi_1 \vee \phi_2 \dri_{\genMet}  (\tss,i) := &  \max\big(\dle \phi_1 \dri_{\genMet}  (\tss,i) , \dle \phi_2 \dri_{\genMet}  (\tss,i) \displaybreak[2]\big)  \\
	\dle \bigcirc \phi \dri_{\genMet}  (\tss,i)  :=  & 
	\displaybreak[2] \left\{ \begin{array}{ll}
	\dle \phi \dri_{\genMet}  (\tss,i+1) & \mbox{ if } i+1\in N\\
	-\infty  & \mbox{ otherwise }  \\
	\end{array} \right.\\
	\dle \phi_1 U_\Ic \phi_2 \dri_{\genMet}  (\tss,i) := & \max_{j \mbox{ s.t. } (t_j-t_i) \in \Ic} \bigg( \min\Big(\dle \phi_2 \dri_{\genMet}  (\tss,j) , \\ & \min_{i \leq k <j} \dle \phi_1 \dri_{\genMet}  (\tss,k) \Big)\bigg)
	\vspace*{-0.2in}
	\end{align*}
	\label{def:mitlrob}
\end{definition}

\hlRev{
The value $\dle \phi\dri_{\genMet} (\tss,0)$ is referred to as the {\it robustness} with which $\tss$ satisfies $\phi$.
For convenience, we just write $\dle \phi\dri_{\genMet} (\tss)$ when $i=0$.
}
As proved in \cite{FainekosP09tcs}, a trace $\tss$ satisfies an STL formula $\phi$ (denoted by $\tss \models\phi$), if $\dle \phi\dri_{\genMet} (\tss)>0$.
On the other hand, a trace $\tss'$ does not satisfy $\phi$ (denoted by $\tss' \not\models\phi$), if $\dle \phi\dri_{\genMet} (\tss')<0$.
An overview of the algorithms that can be used to compute $\robsemD{\varphi}$ is provided in \cite{BartocciEtAl2018survey}.

\begin{example}[Continued from Example \ref{exmp:simple:2car}]
	\label{exmp:simple:spec}
\hlRev{
	Let's assume that we would like to check on the output traces of Fig. \ref{fig:simple:exmp2car:out} the simple safety requirement: {\it after 5 time units, the distance between the adversarial and ego vehicles should always be greater than 0}.
	This requirement is captured by the STL specification $\varphi = \Box_{[5,\infty)} (y^{(a)} - y^{(e)} > 0)$. 
	Then, $\dle \varphi \dri_{\genMet} (\tss) = -1.0841$, which means that the requirement is not satisfied.
	Moreover, the value -1.0841 corresponds to time 5.2446, which is the time of the worst violation of $\phi_1$.  
}	
\end{example}

\subsection{Robustness-Guided Falsification}

The robustness metric can be viewed as a fitness function that indicates the degree to which individual simulations of the system satisfy the requirement~$\varphi$ (positive values indicate that the simulation satisfies $\varphi$). 
Therefore, for a given system $\model$ and a given requirement~$\varphi$, the verification problem is to ensure that for all $\simTr\in \mathcal{L}(\model)$, $\robsemD{\varphi}(\simTr)>0$.

\begin{figure}[t]
	\begin{center}
		\includegraphics[width=0.9\linewidth]{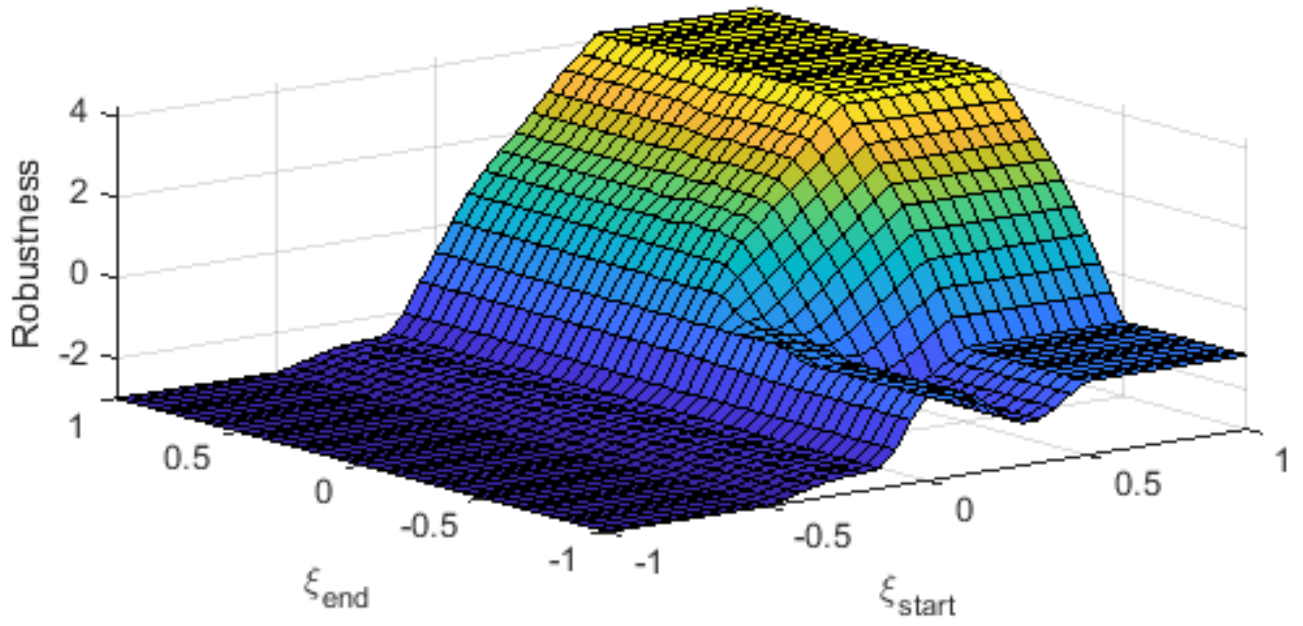} 
	\end{center}
	\vspace{-10pt}
	\caption{The resulting robustness landscape (heatmap) for specification $\varphi$ in Example \ref{exmp:simple:spec}. 
	The initial positions for the two vehicles and the input $\mu$ are fixed. 
	The input signals for $\xi$ are generated by linear interpolation between two input values: $\xi_{start}$ at time 0 and $\xi_{end}$ at time 10.}
	\label{fig:simple:rob:surf}
\end{figure}

Let $\varphi$ be a given STL property that the system is expected to satisfy.  
\hlRev{
The robustness metric $\robsemD{\varphi}$ maps each simulation trace $\simTr$ to a real number $r$ (see Fig. \ref{fig:simple:rob:surf} for an example).  
}
Ideally, for the STL verification problem, we would like to prove that $\inf_{\simTr \in \mathcal{L}(\Sigma)} \dle \varphi \dri_{\genMet} (\simTr) > \varepsilon > 0$ where $\varepsilon$ is a desired robustness threshold.  
\hlRev{
Unfortunately, in general, the problem is not algorithmically solvable \cite{HenzingerKPV98jcss}, that is, there does not exist an algorithm that can solve the problem.
Hence, instead of trying to prove that the property holds on the system, we will try to demonstrate that it does not hold on the system when the system is unsafe.
In other words, we are searching for a trajectory (trace) which {\it falsifies} the requirement (i.e., a trace that demonstrates that the specification is false).
This is the topic of the next section.
}

\subsection{Falsification and Critical System Behaviors} \label{sec:critical}

In this work, we focus on the task of identifying critical system behaviors, 
including falsifying traces. 
%
%
To identify falsifying system behaviors, we leverage existing work on \emph{falsification}, which is the process of identifying system traces $\simTr$ that do not satisfy a given specification $\varphi$.
The STL falsification problem is defined as: 
Find $\simTr\in \mathcal{L}(\Sigma)$ s.t. $\robsemD{\varphi}(\simTr) < 0$.
One successful approach in addressing the falsification problem is to pose it as a global non-linear optimization problem:
\begin{eqnarray} \label{eqn:T_star}
\simTr^\star = \arg \min_{\simTr \in \mathcal{L}(\Sigma)}  \robsemD{\varphi}(\simTr).
\end{eqnarray}
If the global optimization algorithm converges to some local minimizer $\tilde \simTr$ such that $\robsemD{\varphi}(\tilde \simTr)<0$, then a counterexample (adversarial sample) has been identified, which can be used for debugging (or for training).
\hlRev{
Considering the robustness heatmap in Fig. \ref{fig:simple:rob:surf}, any point with robustness below 0 would be a counterexample for the specification of Example \ref{exmp:simple:spec}.
}
In order to solve this non-linear non-convex optimization problem, a number of stochastic search optimization methods can be applied (e.g., \cite{AbbasFSIG13tecs} -- for an overview see \cite{HoxhaEtAl14difts,KapinskiEtAl2016csm}).
We leverage existing falsification methods to identify falsifying examples the autonomous driving system.



\subsection{Covering Arrays}
\label{sec:covering_arrays}

In software systems, there can often be a large number of discrete input parameters that affect the execution path of a program and its outputs.
The possible combinations of input values can grow exponentially with the number of parameters.
Hence, exhaustive testing on the input space becomes impractical for fairly large systems.
A fault in such a system with $k$ parameters may be caused by a specific combination of $t$ parameters, where $1 \leq t \leq k$.
One best-effort approach to testing is to make sure that all combinations of any $t$-sized subset (i.e., all $t$-way combinations) of the inputs are tested.

A \textit{covering array} is a minimal number of test cases such that any $t$-way combination of test parameters exist in the list \cite{hartman2005software}.
Covering arrays are generated using optimization-based algorithms with the goal of minimizing the number of test cases.
We denote a $t$-way covering array on $k$ parameters by $CA(t, k, (v_1, ..., v_k))$, where $v_i$ is the number of possible values for the $i^{th}$ parameter.
The size of the covering array increases with increasing $t$, and it becomes an exhaustive list of all combinations when $t=k$.
Here, $t$ is considered as the \textit{strength} of the covering array.
In practice, $t$ can be chosen such that the generated tests fit into the testing budget.
Empirical studies on real-world examples show that more than $90$ percent of the software failures can be found by testing 2 to 4-way combinations of inputs \cite{kuhn2013introduction}.

\begin{figure}[b]
	\begin{center}
		\begin{tabular}{|c|c|c|c|}
			\hline
			\multicolumn{2}{|c|}{\textbf{}} & \multicolumn{2}{|c|}{\textbf{Color Combinations}}\\ \cline{3-4}
			\multicolumn{2}{|c|}{\textbf{}} & Blue car, blue pants & White car, white pants\\ \hline
			\multirow{4}{0.25cm}{\rotatebox{90}{Vehicle Type}}& \multirow{2}{0.2cm}[1.5cm]{A}  &\includegraphics[width=.35\columnwidth]{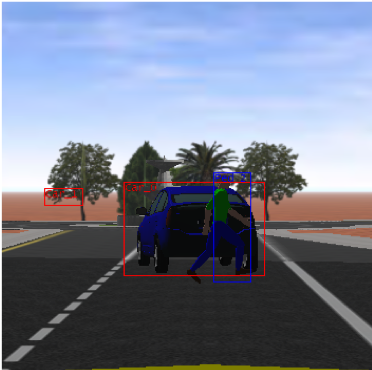}\hspace*{0in}&\includegraphics[width=.35\columnwidth]{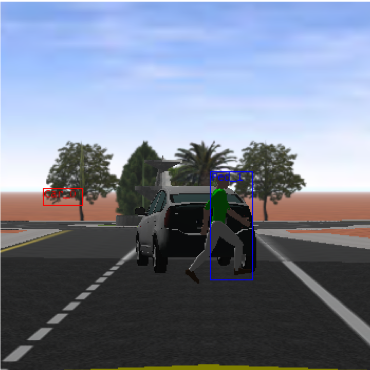}\hspace*{0in}\\
			&&(a) & (b)\\ \cline{2-4}
			&\multirow{2}{0.25cm}[1.5cm]{B} &\includegraphics[width=.35\columnwidth]{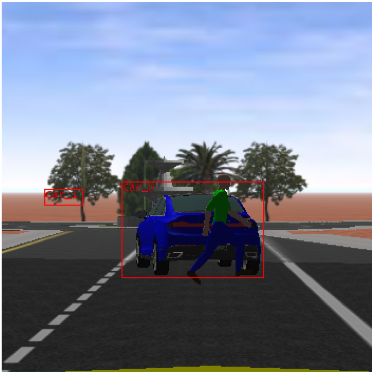}\hspace*{0in}&	\includegraphics[width=.35\columnwidth]{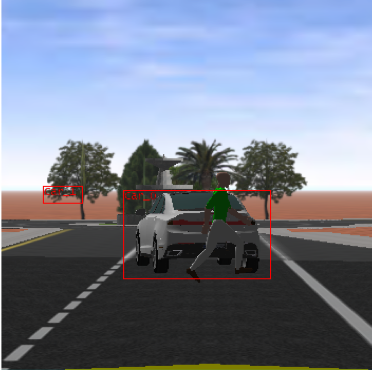}\hspace*{0in}\\
			&&(c) & (d)\\ \hline
		\end{tabular}
		\caption{Specific configurations impacting DNN performance.}
		\label{fig:ca_example}
	\end{center}
\end{figure}

Despite the $t$-way combinatorial coverage guaranteed by covering arrays, a fault in the system possibly may arise as a result of a combination of a number of parameters larger than $t$.
Hence, covering arrays are typically used to supplement additional testing techniques, 
like uniform random sampling (fuzzing).
We consider that because of the nature of the training data or the network structure, NN-based object detection algorithms may be sensitive to a certain combination of properties of the objects in the scene.
Fig. \ref{fig:ca_example} shows outputs of a DNN-based object detection and classification algorithm for 4 different combinations of vehicle type, vehicle color and pedestrian pants color while all other parameters like position and orientation of the objects are the same.
In a comparison between configurations (a) and (b), the vehicle type does not change but the vehicle and pedestrian pants colors change from blue to white. 
While both the car and the pedestrian are detected in configuration (a), the pedestrian is detected but the car is not detected in configuration (b);
however, in a comparison between configurations (b) and (d), if we fix the vehicle and pedestrian pants colors to be white but change the vehicle type, then the car is detected but the pedestrian is not detected.
We can also see that the size of the detection box is different between configurations (c) and (d), for which the vehicle type is the same but the vehicle and pedestrian pants colors are different.
Our observation is that the characterization of the errors is generally not as simple as saying that all white colored cars are not detected. Instead, the errors arise from some combination of subsets of discrete parameters.
Because of this combinatorial aspect of the problem, covering arrays is a good fit to test DNN-based object detection and classification algorithms.
In Sec. \ref{sec:framework}, we describe how \toolname combines covering arrays to explore discrete and discretized parameters with falsification on continuous parameters.

\section{Requirements}
\label{sec:stl_specs_for_experiments}

In this section, we provide five STL requirements intended for the autonomous driving system.
Each requirement is used to target specific aspects of safety and performance.
Also, we describe how analysis results related to each of the requirements can be used to enhance either the controller design or a testing phase of the development process. 

\subsection{STL Requirements}

This section describes each of the requirements that we use in the sequel to evaluate an ADS design with our virtual framework.
We provide these requirements to illustrate how STL can be used to describe four different types of behavior expectations for an ADS: \emph{system-level} safety, \emph{subsystem-level} performance, \emph{subsystem-to-system} safety, and system-level performance (driving comfort) requirements.
In the following, when an object $i$ and a sensor $s$ are clear from the context, we will drop the indices from the notation.

\paragraph*{Requirement $\systemreq$}\label{para:dontcrash}
{\it The ego vehicle should not collide with an object.}

This requirement is an example of a \textit{system-level safety requirement}.
It is used to ensure that the ego vehicle does not collide with any object in the environment. Behaviors that do not satisfy this requirement correspond to unsafe performance by the AV. 
These cases are valuable to identify in simulation, as they can be communicated back to the control designers so that the control algorithms can be improved. 

The formal requirement in STL is:

\[\systemreq_i =\Box (\neg \pi_{i, coll} )\]
where $\pi_{i, coll} = dist(i, ego) < \epsilon_{dist}$.

In the above specification, 
$i$ corresponds to an object in the environment, such as an agent vehicle or a pedestrian.  
$\textit{dist}(i, ego)$ gives the minimum Euclidean distance between the 
boundaries of the Ego vehicle and the boundaries of object~$i$.
The specification basically indicates that the Ego vehicle should not collide with object $i$.

In practice, we consider a separate requirement for each object in the environment and all of them are checked conjunctively, 
i.e., if there are $M_o$ objects, then $\systemreq = \bigwedge_{i=1}^{M_o} \systemreq_i$.

\begin{remark}
\hlRev{
As we have indicated in \cite{TuncaliPF16itsc,TuncaliF19itsc}, $\systemreq$ can be too restrictive and pessimistic for an adversarial testing environment.
Namely, if the requirement simply states ``do not collide with a moving object", then the test engine will attempt to generate agent trajectories that purposefully try to collide with the ego vehicle.
If the adversaries are powerful enough, then the ego vehicle cannot avoid such collision cases.
Hence, depending on the test scenario it may be necessary to enforce  $\systemreq$ only for static objects, or impose additional assumptions formalizing when the ego vehicle is supposed to be able avoid collisions, e.g., \cite{ShalevSS2017arxiv,LoosPN11fm}.
}
\end{remark}

\paragraph*{Requirement $\sensorIreq$}\label{para:sensor1}
{\it Sensor $s$ should detect visible obstacles within $t_1$ time units.}

This requirement is an example of a \textit{subsystem-level requirement}. 
This particular example can be considered as a requirement on the sensor or perception subsystems.
The requirement indicates that the perception system or a specific sensor $s$ should not fail to detect an object for an excessive amount of time, i.e., more than $t_1$ time.

The requirement is as follows.
\begin{align*}
\sensorIreq_{i,s} = \Box \big(&(W(i,s) \wedge \neg D(i,s)) \implies\\
& \Diamond_{[0,t_1]}(D(i,s) \vee \neg W(i,s))\big)
\end{align*}
Here,  $W(i,s)$ denotes that object $i$ is physically \emph{visible} to sensor $s$. For our framework, $s\in \{ CCD, lidar, radar, combined \} $, where $combined$ represents the total perception system, that is, fusion of all available sensors. 
The predicate $D(i,s)$ evaluates to true when sensor $s$ detects object $i$. 
A (non-unique) description of this requirement in natural language is ``it is always true that for any time when object $i$ is visible and not detected by sensor $s$, then there exists a time instant between $0$ and $t_1$ that object $i$ is either detected or it should be invisible to the sensor".

\paragraph*{Requirement $\sensorIIreq$}\label{para:sensor2}
{\it Localization errors should not be too large for too long.}

This requirement is another sensor-level requirement and specifies that the localization of an object that is based on a particular sensor should provide sufficient accuracy, within an adequate time after the object becomes visible to the sensor:
\begin{align*}
\sensorIIreq_{i,s} =& \Box \big((W(i,s) \wedge (\neg D(i,s) \vee E(i,s)>\epsilon_{err})) \implies \\ &\Diamond_{[0,t1]}( \neg W(i,s) \vee (D(i,s) \wedge E(i,s)<\epsilon_{err}) )\big)
\end{align*}
In $\sensorIIreq_{i,s}$, $E(i,s)$ is the difference between object $i$'s location and its location as estimated using information from sensor $s$. 
The constant $\epsilon_{err}$ is a threshold on the acceptable error between the actual position of $i$ and its estimated position.

To understand the requirement, consider the situation where either an object is not detected (i.e., $\neg D(i,s)$) or there is a large error in the localization of the object (i.e., $E(i,s)>\epsilon_{err}$), we refer to this case as \textit{``poor detection"} of the object.
We can interpret the requirement as follows: ``it is always true that whenever  object $i$ is visible to sensor $s$ and is poorly detected by sensor $s$, then there exists an instant, within a time period from $0$ to $t1$, that either object $i$ is invisible to sensor $s$ or the object is detected and the localization error is small, as computed using information from sensor $s$".
This requirement basically limits the amount of time the sensor error can be greater than a given threshold.

\paragraph*{Requirement $\sensetosystemreq$}\label{para:sensortosystem}
{\it A sensor-related fault should not lead to a system-level fault.}

This is an example of a \textit{subsystem-to-system requirement}.
This requirement relates sensor-level behaviors to system-level behaviors. 
The purpose is to isolate behaviors where a sensor fault results in a collision.
The expectation is that the system as a whole should be robust to failure of a single sensor:


%
\begin{align*}
\sensetosystemreq_{i,s} =&\Box \neg \Big( \Box_{[0,t1]}  \big( \neg \pi_{i, coll} \wedge W(i,s) \wedge \\ 
& (\neg D(i,s) \vee E(i,s) > \epsilon_{err} ) \big)  \wedge \Diamond_{(t1,t2]}  \pi_{i, coll}   \Big)
\end{align*}

The above requirement designates that there should not be a period of $t1$ where a visible object is not accurately detected and no collision occurs, followed immediately by a period of length $t2-t1$ that contains a collision.
In other words, the requirement indicates that 
a system level fault (collision) should not occur within a short time after a sensor fault.
A behavior that violates this requirement does not necessarily indicate that the sensor fault \emph{caused} the system fault, but it suggests a correlation, as it points to a behavior wherein 
the system fault occurs a short time after the sensor fault.
Providing behavior examples that violate this requirement can help to pinpoint the cause of system-level faults.


\paragraph*{Requirement $\perfreq$}\label{para:dontbrake}
{\it The vehicle should not do excessive braking unnecessarily or too often.}

This is a \textit{system-level performance (driving comfort)} requirement, in that it requires that the system not brake unnecessarily or too often, thereby causing discomfort for the passengers:

\begin{align*}
\perfreq = \Box \Big(& \neg \Box_{[0,t1]}(B\wedge\neg \collestimated)  \wedge \\ &\neg\big( \edge \wedge \Diamond_{(0,t2]}( \edge \wedge \Diamond_{(0,t2]}\edge ) \big) \Big),
\end{align*}
%
Here, $\collestimated$ is a variable that is true when the Ego vehicle is estimated to collide in the future with another object in the environment, based on a simplified model of future behaviors.
The simplified model that we use for future trajectory estimation is the Constant Turn Rate and Velocity (CTRV) model \cite{Schubert2008}.
The proposition $B$ represents that the amount of braking force applied by the controller exceeds half of the available braking force.
Finally, $\edge = B \wedge \bigcirc \neg B$ represents the event of releasing the brake, i.e., a $true$ value of $B$ followed by a $false$ value in the next sample.

To understand the meaning of requirement $\perfreq$, consider the following part of the requirement:
\[\Box \Big(\neg \Box_{[0,t1]}(B\wedge\neg \collestimated)\Big),\] 
which requires that the system not apply excessive braking for more than a specific amount of time $(t1)$ while there is no collision predicted. This essentially stipulates that the system should not unnecessarily brake for a prolonged amount of time. 
Next, consider the second part of requirement $\perfreq$:
\[ \Box \Big(\neg\big( \edge \wedge \Diamond_{(0,t2]}( \edge \wedge \Diamond_{(0,t2]}\edge ) \big) \Big),\] 
which indicates that there should not be an ``on-off" behavior, followed by another ``on-off" behavior, followed by a third ``on-off" behavior, with less than $t2$ between each other. 
This essentially requires that the brakes not be applied and released too often.
Thus, this is a riding comfort requirement.

\subsection{Development Process Support}

We describe how requirements $\systemreq$ through $\perfreq$ can be used to support both the controller design and testing phases of the development process.
For all of the requirements, any detected
violation (falsification) should be linked back to the
conditions that caused the violation.

\begin{figure}[htp]
	\begin{center}
			\includegraphics[width=8.5cm]{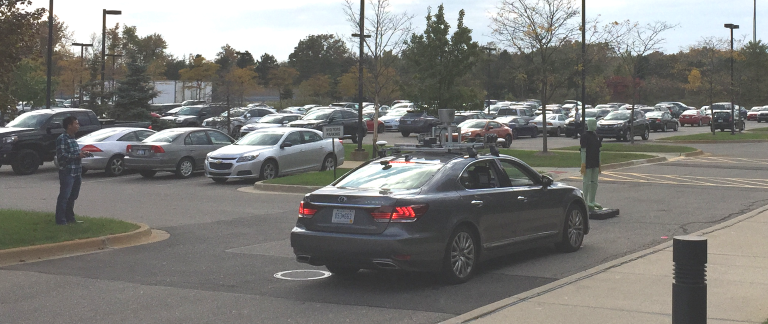}
		\caption{Robotic pedestrian surrogate target with a Toyota autonomous vehicle.}
		\label{fig:ToyotaDummy}
	\end{center}
\end{figure}

Consider the first requirement, $\systemreq$, ``The vehicle should not
collide with an object": if the vehicle does collide with an object,
then we would go back and see what conditions caused such an event,
for example, whether the vehicle speed trace exhibited an anomaly or
whether the
vehicle was moving erratically.  Testing for collision
avoidance is well established in the field of ADAS.  Often
inflatable and other destructible targets are employed; 
for example,  see \cite{LeBlanc2013} and Fig. \ref{fig:ToyotaDummy}.

In the requirement ``Sensor should detect visible obstacles", we
focus on the detection of an obstacle as operational imperative.  If
the sensor fails to detect the object within a time interval, then the
requirement is violated. This is essentially a sensor-level
requirement (visible but not detected), and test engineers can set a
real-world experiment to verify it relatively easy because it is
decoupled from others (one-term inequality, sensor by sensor).

The requirement "Localization error should not be too high for too
long" is important to verify (falsify) for both ego-location and
position identification of other agents in the environment.
Placing an ego-vehicle in the correct pose on the road is usually
not achieved by simply relying on GPS signal processing, due to the
GPS tendency to ``jump" unpredictably, but instead by estimating and
dynamically refining the pose through landmark observations, such as
road edges, vertical elements such as light poles, and signs.  Assuming that the ego-vehicle
localization is done with sufficient accuracy, the remaining task of
localization is to make sure that the location of other agents, especially those in
the planned path of the ego vehicle, are estimated with
sufficient accuracy.  Often a grid-based representation centered on
the ego-vehicle is employed (e.g., \cite{Petrovskaya2009}).
Estimating $E(i,s)$ in $\sensorIIreq$ is not trivial, but 
practical approaches exist that can be used by test engineers (e.g., \cite{Grabe2009}).

The requirement ``A sensor-related fault should not lead to a
system-level fault" is a form of robustness requirement.
This is similar to a requirement that the system should have 
no ``single point of failure", which enforces that  
the
failure of any single component will not cause the system to fail
(for example, see \cite{ISO2018}). We make an important clarification 
which is practical but limiting in scope: no failure should occur
within the specified (short) time after the fault.  Test engineers
could readily use examples of behavior provided in the course of
falsifying this requirement.

Lastly, the requirement ``The vehicle should not brake
too often" is an example of a possible set of requirements
designed to establish how comfortable the ride in the vehicle is.  It
is known that autonomous vehicles could induce motion sickness in
passengers if the vehicle control system does not comply with human
physiology \cite{Elbanhawi2015}, \cite{Green2016}.  A
better requirement may well be developed using fuzzy set theory and
further refined for a specific target group of passengers (e.g.,
elderly people). An alternative requirement could be defined by counting
the number of occurrences of an event within a total time period, instead 
of relating one occurrence to another. Such a requirement can be defined as a 
Timed Propositional Temporal Logic (TPTL) specification. 
TPTL is a generalization of STL which is also supported in our framework \cite{dokhanchi2016efficient}.

\section{\toolname Framework} \label{sec:framework}

We describe \toolname, a framework for performing testing and analysis of autonomous driving systems in a virtual environment.
\hlRev{
The framework is publicly available as an add-on to \staliro \cite{AnnapureddyLFS11tacas}:
}

\begin{center}
\hlRev{
 \url{https://sites.google.com/a/asu.edu/s-taliro/s-taliro/sim-atav}
}
\end{center}

\noindent The simulation environment used in \toolname is based on the open source simulator Webots \cite{michel2004cyberbotics} and includes a vehicle perception system, a vehicle controller, and a model of the physical environment.
The perception system processes data from three sensor systems: CCD camera images, lidar, and radar.
The framework uses freely available and low cost tools and can be run on a standard desktop PC.
Later, we demonstrate how \toolname can be used to implement traditional testing approaches, as well as advanced automated testing approaches that were not previously possible using existing frameworks.

\begin{figure}[]
\begin{centering}
\includegraphics[width=0.95\columnwidth,trim={0 0.75cm 2.5cm 0},clip]{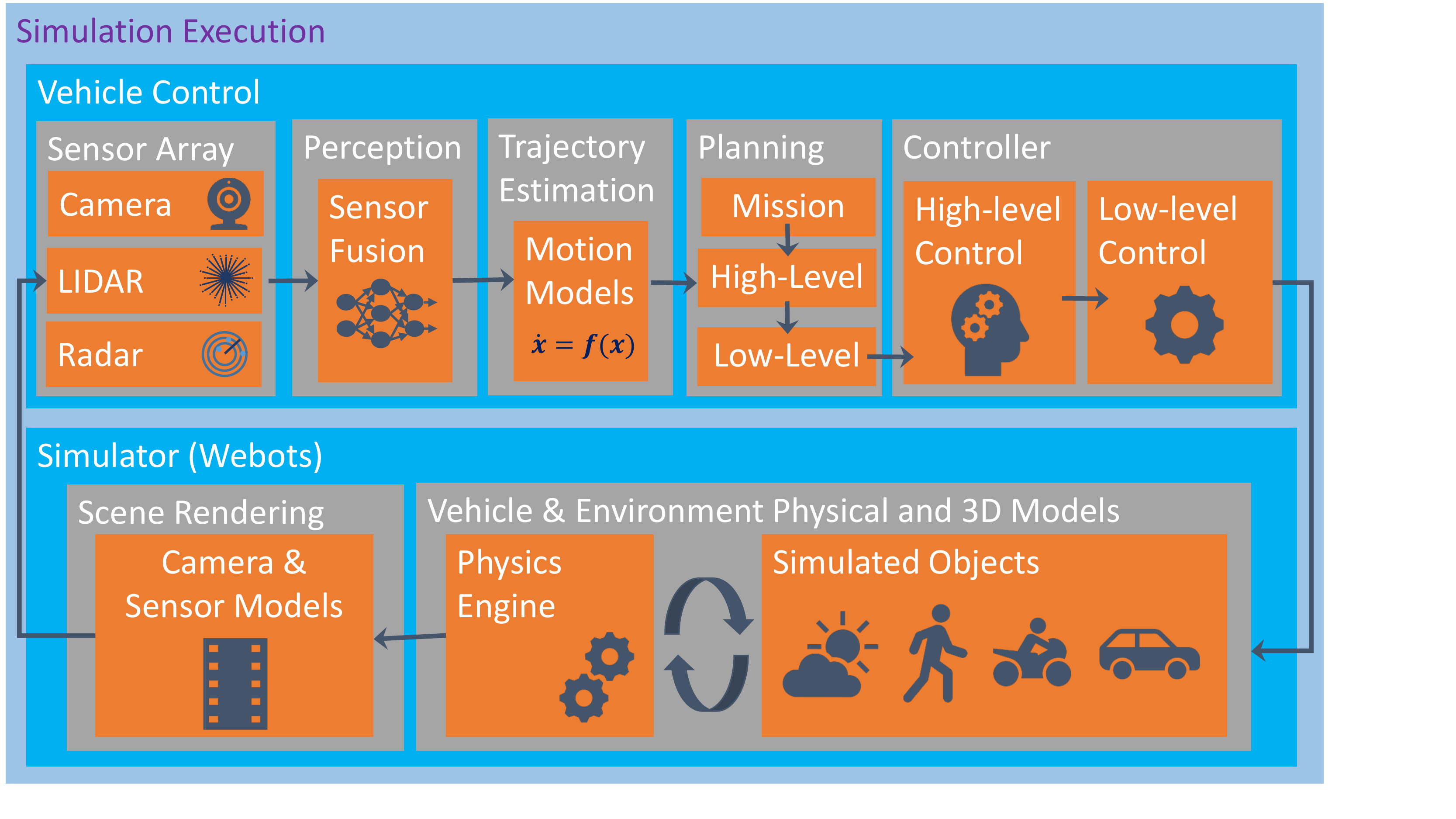}
\caption{Overview of the simulation environment. \label{fig:framework}}
\end{centering}
\end{figure}

Fig. \ref{fig:framework} shows an overview of the simulation environment.
The environment consists of a Simulator and a Vehicle Control system.
The Simulator contains models of the ego vehicle, agents, and other objects in the environment (e.g., roads, buildings).  
The Simulator outputs sensor data to the Vehicle Control system.
The sensor data includes representations of CCD camera, lidar, and radar data.
Simple models of the sensors are used to produce the sensor data. For example, synthetic CCD camera images are rendered by the Simulation system, as if they came from a camera mounted on the front of the ego vehicle.
The Vehicle Control system contains models of the Perception System, which performs sensor data processing and sensor fusion.
The Controller uses the output of the Perception System to make decisions about how to actuate the AV system. 
Actuation commands are sent from the Controller to the Simulator.

Simulations proceed iteratively. At each instant, sensor data is processed by the Vehicle Control, which then makes an actuation decision. The actuation decision is then transmitted back to the Simulator, which uses the actuation commands to update the physics for the next time instant. This process is repeated until a designated time limit has been reached.

The Vehicle Control system is implemented in Python.
We use simplified algorithms to implement the subsystems of the vehicle control, which 
is sufficient in this case, as the purpose of this investigation is to evaluate new \emph{testing} methodologies and not to evaluate a real AV control design;
however, we note that it is straightforward to replace our algorithms with production versions to test real control designs.

\begin{figure}[h]
	\begin{center}
		\includegraphics[width=\columnwidth]{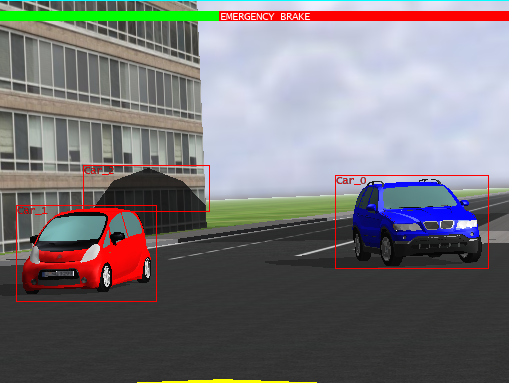}
		\caption{Outputs from the SqueezeDet DNN, based on a synthesized camera image.}
		\label{fig:dnn_camera}
	\end{center}
\end{figure}

To process CCD image data, we use a lightweight DNN, SqueezeDet, which performs object detection and classification~\cite{wu2017squeezedet}.
SqueezeDet is implemented in TensorFlow\texttrademark \cite{abadi2016tensorflow}, and it outputs a list of object detection boxes with corresponding class probabilities.
This network was originally trained on real image data from the KITTI dataset \cite{geiger2012we} to achieve accuracy comparable to the popular AlexNet classifier \cite{krizhevsky2012imagenet}.
We further train this network on the virtual images generated in our framework.
Fig. \ref{fig:dnn_camera} shows an example output from SqueezeDet, based on a synthetic image produced by our simulator.
The image shows two vehicles correctly detected and classified, along with a portion of a shadow that is incorrectly classified as a vehicle.

\hlRev{
To process lidar point cloud data, we first cluster the received points based on their positions using the DBSCAN algorithm \cite{ester1996density,schubert2017dbscan}. 
Then, we estimate the existence and types of the objects based on how well the dimensions of the clusters match with the dimensions of expected object types such as pedestrians or cars.
For estimating the object type in the received radar targets, we use the radar signal power.
}
We implement a simple sensor fusion algorithm that relates and merges the object detections from camera, lidar, and radar with a simple logic.
\hlRev{
Our sensor fusion algorithm makes some rule-based decisions, such as if the object type can be recognized by the camera, discard the type estimations done by the lidar and radar, and on the other hand if radar or lidar is able to detect to position of the object, discard the position estimations computed by the camera.
}
It also utilizes the expected current positions of previously detected objects.
\hlRev{
A simple implementation of an unscented Kalman filter is used to estimate the current and future trajectories of the objects using the CTRV (Constant Turn Rate and Velocity) model \cite{wan2000unscented,Schubert2008}.
}

\begin{figure}[t]
	\begin{center}
		\includegraphics[width=0.9\columnwidth]{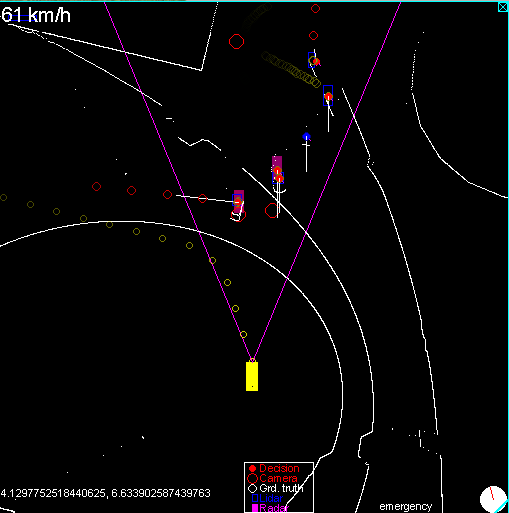}
		\caption{Sensor fusion outputs.}
		\label{fig:sensor_fusion}
	\end{center}
\end{figure}

Fig. \ref{fig:sensor_fusion} illustrates outputs from the sensor fusion system.
In the figure, the solid yellow box in the middle represents the Ego vehicle.
Yellow circles in front of the ego vehicle represent the estimated future trajectory of the Ego vehicle.
Small white dots represent lidar point cloud data.
The colored dots and rectangles represent detected objects, with their estimated orientation indicated with a white line in front of them.
Expected future positions of agent vehicles with respect to the ego vehicle are represented by red circles.

Our simple planner receives as inputs the high level target path and target speed, and the outputs of the sensor fusion and trajectory estimation modules.
It assigns collision risk level to the target objects with a simple logic and outputs the risk assessments and a target speed, which depends on the target speed of the mission or other factors, such as the distance to a sharp turn ahead.

Our control algorithm implements simple path and speed tracking and \emph{collision avoidance} features. 
The controller receives the outputs of the planner.
When there is no collision risk, the controller drives the car with the target speed and on the target path.
When a future collision with an object is predicted, it applies the brakes at a level proportional with the risk assigned to the object.

The environment modeling framework is implemented in Webots \cite{michel2004cyberbotics}, a robotic simulation framework that models the physical behavior of robotic components, such as manipulators and wheeled robots, and can be configured to model autonomous driving scenarios.
In addition to modeling the physics, a graphics engine is used to produce images of the scenarios.
In \toolname, the images rendered by Webots are configured to correspond to the image data captured from a virtual camera that is attached to the front of a vehicle.

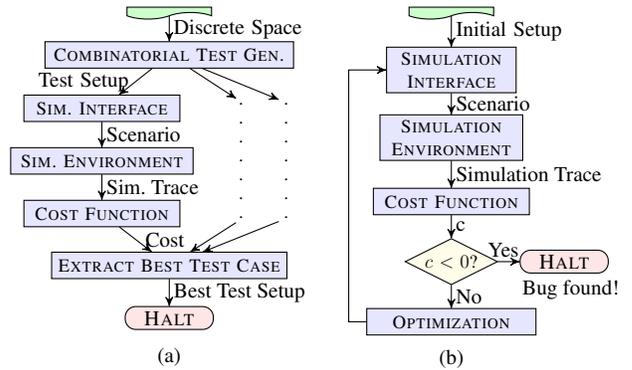
\begin{figure}[]
	\begin{centering}
		\begin{tikzpicture}[>=stealth',scale=0.5, every node/.style={scale=0.5}]

\tikzset{
  font={\fontsize{16pt}{16}\selectfont}}

\tikzstyle{stext}=[font=\fontsize{14}{14}\selectfont]
\tikzstyle{smalltext}=[font=\fontsize{14}{14}\selectfont]
\tikzstyle{block}=[draw,fill=white,rectangle,minimum height=2em,text
                   width=4.5em,fill=blue!10,stext,inner sep=4pt,align=center,execute at begin node=\setlength{\baselineskip}{1.2em}]
\tikzstyle{file}=[tape,fill=red!10,tape bend top=none,draw,inner
                  sep=2pt,text width=5em,align=center,smalltext,execute at begin node=\setlength{\baselineskip}{1.2em}]
\tikzstyle{note} = [coordinate]

\node[file,fill=green!20, text width=6em]  (init setup) {};

\node[block,below of= init setup, node distance=16mm, text width=9em] (sim int) {\textsc{Simulation \\Interface}};

\node[block,below of=sim int, node distance=18mm, text width=10em] (s) {\textsc{Simulation \\Environment}};

\node[block,below of=s,node distance=17mm,text width=11em] (cost) {\textsc{Cost Function}};

\node[draw,smalltext,diamond,aspect=2,text width=5em,below of=cost,fill=yellow!10,node distance=16mm,inner sep=0pt,align=center] (decision) 
     {$c<0$?};

\node[block,below of=decision,node distance=16mm,text width=12em] (optim) {\textsc{Optimization}};

\node[draw,rounded corners,align=center, right of= decision, node distance=30mm,fill=red!10,text width=6em] (stop) 
     {\textsc{Halt}};
\node [note,name=bugfound, font=\Large, below right=0.4cm and -0.5cm of stop,label={Bug found!}] {};

\draw[->] (init setup) -- node[name=initalize, right, text width=9em,execute at begin node=\setlength{\baselineskip}{1.2em}] {Initial Setup} (sim int);

\draw[->] (sim int) -- node[name=initalize, right, text width=5em,execute at begin node=\setlength{\baselineskip}{1.2em}] {Scenario} (s);

\draw[->] (s) -- node[name=simresults, right]{Simulation Trace} (cost);

\draw[->] (cost) -- node[name=costout, right]{c} (decision);

\draw[->] (decision) -- node[name=no,right]{No} (optim);

\draw[->] (decision) -- node[name=yes,above, near start]{Yes} (stop);

\draw [->] (optim.west) -| node[name=newparam, left, near end, text width=3em,execute at begin node=\setlength{\baselineskip}{1.2em}]{} ($(sim int.west)+(-1.0,0.0)$) -- (sim int.west) ;

\node[file, left of=init setup, node distance = 75mm, fill=green!20, text width=6em]  (discrete init) {};
\node[block, below of=discrete init, node distance=12mm, text width=18em] (combinatorial) {\textsc{Combinatorial Test Gen.}};
\draw[->] (discrete init) -- node[name=discrete space, right, text width=12em,execute at begin node=\setlength{\baselineskip}{1.2em}] {Discrete Space} (combinatorial);

\node[block, below left=1em and -18mm of combinatorial, text width=11em] (sim int discrete) {\textsc{Sim. Interface}};
\draw[->] (combinatorial) -- node[name=setup discrete, left, text width=7em,execute at begin node=\setlength{\baselineskip}{1.2em}] {Test Setup} (sim int discrete);

\node[stext, below right=4mm and -8mm of combinatorial, text width=1em] (s1) {\textsc{.}};
\draw[->] (combinatorial) -- node[name=setup discrete 2, left, text width=7em,execute at begin node=\setlength{\baselineskip}{1.2em}] { } (s1);
\node[stext, below of=s1, node distance=5mm, text width=1em] (s3) {\textsc{.}};
\node[stext, below of=s3, node distance=5mm, text width=1em] (s4) {\textsc{.}};
\node[stext, below of=s4, node distance=5mm, text width=1em] (s5) {\textsc{.}};
\node[stext, below of=s5, node distance=5mm, text width=1em] (s6) {\textsc{.}};
\node[stext, below of=s6, node distance=5mm, text width=1em] (s7) {\textsc{.}};
\node[stext, below of=s7, node distance=5mm, text width=1em] (s11) {\textsc{.}};

\node[stext, right of=s1, node distance=12mm, text width=1em] (s2) {\textsc{.}};
\draw[->] (combinatorial) -- node[name=setup discrete 2, left, text width=7em,execute at begin node=\setlength{\baselineskip}{1.2em}] { } (s2);
\node[stext, below of=s2, node distance=5mm, text width=1em] (s3) {\textsc{.}};
\node[stext, below of=s3, node distance=5mm, text width=1em] (s4) {\textsc{.}};
\node[stext, below of=s4, node distance=5mm, text width=1em] (s5) {\textsc{.}};
\node[stext, below of=s5, node distance=5mm, text width=1em] (s6) {\textsc{.}};
\node[stext, below of=s6, node distance=5mm, text width=1em] (s7) {\textsc{.}};
\node[stext, below of=s7, node distance=5mm, text width=1em] (s12) {\textsc{.}};

\node[block, below of=sim int discrete, node distance=14mm, text width=13em] (sim env discrete) {\textsc{Sim. Environment}};
\draw[->] (sim int discrete) -- node[name=setup discrete, right, text width=16em,execute at begin node=\setlength{\baselineskip}{1.2em}] {Scenario} (sim env discrete);

\node[block, below of=sim env discrete, node distance=14mm, text width=11em] (cost discrete) {\textsc{Cost Function}};
\draw[->] (sim env discrete) -- node[name=simresults discrete, right]{Sim. Trace} (cost discrete);

\node[block, below of=combinatorial, node distance=56mm, text width=17em] (minimum discrete) {\textsc{Extract Best Test Case}};
\draw[->] (cost discrete) -- node[name=simresults discrete, right]{\ Cost} (minimum discrete);
\draw[->] (s11) -- node[name=simresults discrete, right]{} (minimum discrete);
\draw[->] (s12) -- node[name=simresults discrete, right]{} (minimum discrete);

\node[draw, rounded corners ,align=center, below of=minimum discrete, node distance=14mm,fill=red!10,text width=6em] (stop discrete) 
{\textsc{Halt}};
\draw[->] (minimum discrete) -- node[name=best result discrete, right]{Best Test Setup} (stop discrete);

\node [note,name=alabel, font=\Large, below =0.55cm of stop discrete,label={(a)}] {};
\node [note,name=blabel, font=\Large, below =0.5cm of optim,label={(b)}] {};

\end{tikzpicture}
		\vspace{-0.2in}
		\caption{Flowcharts illustrating the combinatorial testing (a) and falsification (b) approaches. \label{fig:flowchart}}
	\end{centering}
\end{figure}

The process used by \toolname for test generation and execution for discrete and discretized continuous parameters is illustrated by the flowchart shown in Fig. \ref{fig:flowchart}-(a).
\toolname first generates test cases that correspond to scenarios defined in the simulation environment using covering arrays as a combinatorial test generation approach.
The scenario setup is communicated to the simulation interface using TCP/IP sockets.
After a simulation is executed, the corresponding simulation trace is received via socket communication and evaluated using a cost function.
Among all discrete test cases, the most promising one is used as the initial test case for the falsification process shown in Fig. \ref{fig:flowchart}-(b).
For falsification, the result obtained from the cost function is used in an optimization setting to generate the next scenario to be simulated.
For this purpose, we used \staliro \cite{AnnapureddyLFS11tacas}, which is a \matlab toolbox for falsification of CPSs.
Similar tools, such as Breach \cite{donze10cav}, can also be used in our framework for the same purpose.

\section{Testing Application} \label{sec:applications}
In this section, we present an evaluation of our \toolname framework using three separate driving scenarios.
The scenarios are selected to be both challenging for the autonomous driving system and also analogous to plausible driving scenarios experienced in real-world situations.
In general, the system designers will need to identify crucial driving scenarios, based on intuition about challenging situations, from the perspective of the autonomous vehicle control system. 
A thorough simulation-based testing approach will include a wide array of scenarios that exemplify critical driving situations.

For each of the following scenarios, we consider a subset of the requirements presented in Sec. \ref{sec:stl_specs_for_experiments} and describe how to use the results to enhance the development process. We conclude the section with a summary of the results.

\subsection*{Scenario 1}

\begin{figure}[]
	\begin{centering}
		\includegraphics[trim={0 6.5in 6.5in 0.25in}, clip,width=0.75\columnwidth]{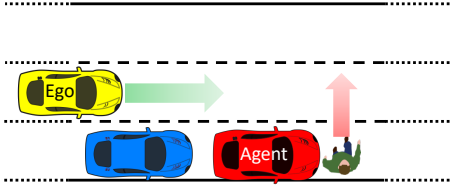}
		\caption{Overview of the scenario 1.}
		\label{fig:scenario1overview}
	\end{centering}
\end{figure}

The first scene that we consider is a straightaway section of a two-lane road, as illustrated in Fig. \ref{fig:scenario1overview}.
Several cars are parked on the right-hand side of the road, and a pedestrian is jay-walking in front of one of the cars, passing in front of the Ego car from right to left. We call this driving scenario model $\modela$.
The scenario simulates a similar setup to the Euro NCAP Vulnerable Road User (VRU) protection test protocols \cite{NCAP2018}.

Several aspects of the driving scenario are parameterized, meaning that their values are fixed for any given simulation by appropriately selecting the model parameters.
The parameters and initial conditions that we use for this scenario are:
\hlRev{
\begin{itemize}
\item Initial speed of the Ego vehicle: $[10, 30] m/s$;
\item Lateral position of Ego w.r.t its lane center: $[-0.8, 0.8] m$;
\item Walking speed of the pedestrian: $[1.5, 6] m/s$;
\item The model of Agent car, which is next to the pedestrian: \textit{from 5 different vehicle models};
\item R, G, B values for the colors of Agent car: $[0,1]$;
\item R, G, B values for the pedestrian's shirt and pants: $[0,1]$.
\end{itemize}
}
We chose the parameterized aspects of the scenarios such that their specific combinations would be challenging to a DNN-based pedestrian detection system that relies on CCD camera images, \textit{e.g., some combinations of agent vehicle shape and color with pedestrian clothing colors may be challenging to an object detection DNN}.
We also chose some of the parameter ranges, \textit{i.e, the vehicle and pedestrian speeds,} so that the scenario is physically challenging for the brake performance.
\hlRev{
The Ego vehicle is longitudinally placed such that the time-to-collision with the pedestrian will be $1.4s$ when the pedestrian is exactly in front of the Ego vehicle (influenced by the Euro NCAP, VRU protection test protocols \cite{NCAP2018}).
}

We evaluate Model $\modela$ against three of the requirements from Sec. \ref{sec:stl_specs_for_experiments}: $\systemreq$, $\sensorIreq$, and $\sensetosystemreq$. These include the system-level requirement, the sensor-level requirements, and the sensor-to-system-level requirement.
We use this collection of requirements for  Model $\modela$  to demonstrate how we can automatically identify each type of behavior using our framework.  

\subsection*{Scenario 2}

\begin{figure}[]
	\begin{centering}
		\includegraphics[trim={2in 0.4in 2.5in 0.8in}, clip,width=\columnwidth]{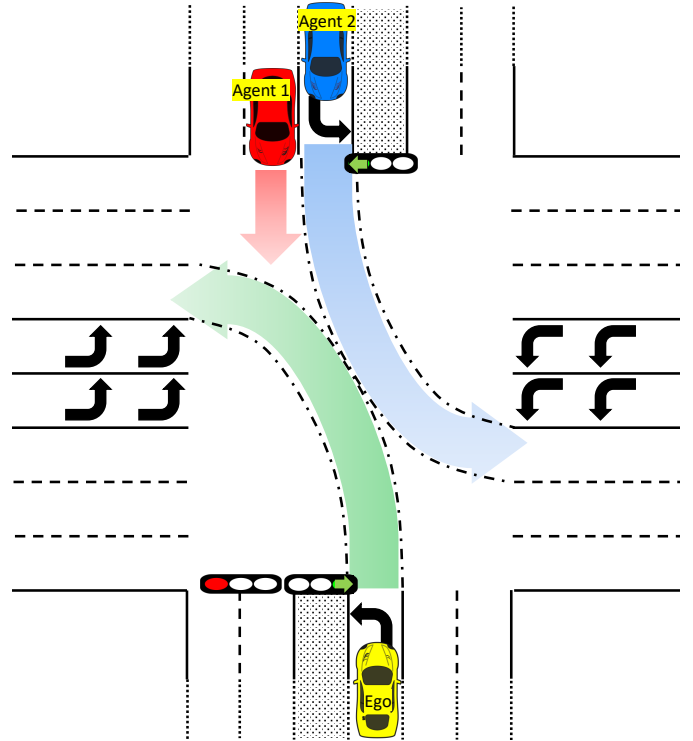}
		\caption{Overview of the scenario 2.}
		\label{fig:scenario2overview}
	\end{centering}
\end{figure}

The next scenario involves a left turn maneuver by the ego vehicle in a controlled intersection, as illustrated in Fig. \ref{fig:scenario2overview}.
 An agent vehicle \textit{(Agent 1)} in the opposing lane unexpectedly passes through the intersection, against a red light, potentially causing a collision with the Ego vehicle.
There is also another agent car \textit{(Agent 2)}, which is making a legal left turn from the opposing lane.
It is incumbent on the Ego vehicle to take action to avoid colliding with the agent vehicles. We call the model of this scenario $\modelb$.

For this experiment, we choose parameters such that the position of Agent 2, or trajectory followed by Agent 1, in combination with the behavior of the Ego, may result in poor performance from the sensor processing or trajectory estimation systems.
For this scenario the search space is:
\hlRev{
\begin{itemize}
	\item Ego vehicle initial speed: $[20, 30] m/s$;
	\item Ego vehicle initial distance to the intersection: $[80, 160] m$;
	\item Agent 1 initial distance to the intersection: $[50, 100] m$;
	\item Agent 1 target speed (initial, when approaching the intersection, when inside the intersection): $[10, 35] m/s$;
	\item Agent 1 lateral position w.r.t its lane center (initial, when approaching the intersection, inside the intersection): $[-0.75, 0.75] m$;
	\item Agent 2 lateral position w.r.t its lane center: $[-0.75, 0.75] m$;
	\item Agent 2 speed: $[3, 15] m/s$;
	\item Agent 2 initial distance to the intersection: $[50, 100] m$.
\end{itemize}
}
We evaluate Model $\modelb$ against requirement $\sensetosystemreq$. The idea in using the sensor-to-system-level requirement is that it is relatively easy, in general, to find behaviors that result in a collision for Model $\modelb$, but many collision cases are not interesting for the designers. This could be because, for example, the agent car is moving too quickly for the ego vehicle to avoid. 
This would be a behavior that is not necessarily caused by any specific incorrect behavior on the part of the ego vehicle. 
Instead, we use $\sensetosystemreq$ to identify behaviors where there is a collision that is directly correlated to  unacceptable performance from the sensor processing system.
These are cases where the sensor data processing or future trajectory estimation system is at fault for the collision. 
Such cases can be easily used to debug specific aspects of the ego vehicle control algorithms.

\subsection*{Scenario 3}

In this last scenario, the ego vehicle is making a left turn through an intersection, while an agent vehicle in the opposing lane is also making a left turn. 
This scene is similar to the Scenario 2, as depicted in Fig. \ref{fig:scenario2overview}, except that Agent 1 is not present in this scenario, only Agent 2, which we refer to as the agent vehicle for this scenario.
If both ego and agent vehicles are not accurately regulating their trajectories during this maneuver, a collision may occur.
We call the model of this scenario $\modelc$.

In this scenario, we search over target trajectories of the ego and agent vehicles.
Below are the parameters that we use:
\hlRev{
\begin{itemize}
	\item Ego vehicle initial speed: $[15, 30] m/s$;
	\item Ego target lateral position w.r.t its lane center when entering the intersection: $[-1,1] m$, longitudinal starting position of the left turn w.r.t. intersection entry point: $[-1, 10] m$, target lateral position w.r.t its lane center when exiting the the intersection: $[-1,1] m$, end position of the left turn w.r.t. intersection exit point: $[-1, 10] m$;
	\item Agent vehicle speed: $[15, 30] m/s$, target lateral position w.r.t its lane center when entering the intersection: $[-1,1] m$, longitudinal starting position of the left turn w.r.t. intersection entry point: $[-1, 10] m$, target lateral position w.r.t its lane center when exiting the the intersection: $[-1,1] m$, and the end position of the left turn w.r.t. intersection exit point: $[-1, 10] m$.
\end{itemize}
}
We evaluate Model $\modelc$ against requirement $\perfreq$. The purpose of considering the performance requirement $\perfreq$ in this case, is that scenario $\modelc$ is difficult to falsify. That is, due to the specific parameter ranges selected for the scenario, it is unlikely that the ego vehicle will collide with the agent vehicle. Instead, in this case, we are interested in identifying situations where the emergency braking system unnecessarily decelerates the ego vehicle, causing unacceptable performance, from a ride-quality perspective.
The scenario can easily lead to unnecessary braking, as the ego and agent vehicles momentarily move toward each other during their left turn maneuvers, which can cause the emergency braking algorithm to decide, incorrectly, that a collision is imminent.
Such 
cases can be a useful feedback to designers, as they can highlight behaviors that are too conservative at the expense of ride quality.

\subsection*{Summary of Test Results}

\begin{table*}[t] 
\centering 
\renewcommand{\arraystretch}{1.5} 
\begin{tabular}{  l | c | c | c | c | c  }
& \multicolumn{3}{|c|}{\bf{Model $\modela$}} & \bf{Model $\modelb$} & \bf{Model $\modelc$} \\
\hline
\bf{Requirement} & $R1$ & $R2$ & $R4$ & $R4$ & $R5$ \\
\hline
\multirow{2}{1.5cm}{\bf{Testing Modality}} & \multirow{2}{1.5cm}{CA+ Falsification} & \multirow{2}{1.5cm}{CA+ Falsification} & \multirow{2}{1.5cm}{CA+ Falsification} & \multirow{2}{1.5cm}{Falsification} & \multirow{2}{1.5cm}{Falsification} \\
&&&&& \\ 
\hline
\bf{Active Sensors} & CCD & CCD & CCD & CCD, Radar, LIDAR & CCD, LIDAR \\
\hline
\multirow{2}{1.5cm}{\bf{Computation Time}} & \multicolumn{3}{|c|}{CA: 2h, 10min.} &\multirow{2}{*}{2h, 3min.}& \multirow{2}{*}{9h, 40min.}\\ \cline{2-4}
&Fals.:3h, 33min.&3h, 35min.&3h 34min. &&\\
\hline
\multirow{2}{1.5cm}{\bf{No. Simulations}} & \multicolumn{3}{|c|}{CA: 195} &\multirow{2}{*}{58}& \multirow{2}{*}{232}\\ \cline{2-4}
&Fals.:300&300&300&& \\
\hline
\multirow{2}{1.5cm}{\bf{Falsification Obtained}} &67 by CA + & 65 by CA + & 67 by CA +& \multirow{2}{*}{\checkmark} & \multirow{2}{*}{\checkmark} \\
&5 by falsification&8 by falsification&12 by falsification && \\
\hline
\bf{\begin{minipage}[t]{1.5cm} \raggedright Application of Results\end{minipage}} & 
\multicolumn{3}{|c|}{Lowest robustness cases used to create critical tests.} & \begin{minipage}[t]{3.9cm} \raggedright
Falsifying cases relate to processing of specific sensor; aids in controller design improvement. \end{minipage} &\begin{minipage}[t]{3.75cm} \raggedright Poor performance cases used to improve controller design in modeling phase. \end{minipage}\\
\hline
\end{tabular} 
\caption{Results from autonomous driving tests using virtual framework.}
\label{table:experiment_results}
\end{table*}
We present results from experiments demonstrating the application of our framework to the scenarios and requirements described above.
Table~\ref{table:experiment_results} summarizes the results.
For each case study, Table \ref{table:experiment_results} indicates the requirements used to test each model, the testing approach used, the set of active sensors used, and a summary of the results.
We discuss the test generation outcomes in detail below.

\subsubsection*{Covering array and falsification on Model $M_1$}
In our previous work \cite{Tuncali2018}, we proposed and studied the effectiveness of a testing approach that first uses covering arrays to discover critical regions, based on a set of discrete parameters, then uses those results as the initial points for robustness guided falsification.
Here, we apply that approach on model $M_1$ for 3 different requirements, $R1$, $R2$ and $R4$.
In model $M_1$, we focus on the camera sensor and DNN-based object detection and classification algorithm. 
\hlRev{
Because of this, most of our parameters are colors of pedestrian clothing and the agent vehicle, to which an object detection system may be sensitive to as described in Sec. \ref{sec:stl_specs_for_experiments}.
}

\hlRev{
We create a mixed-strength covering array from a discretization of the parameterized variables.
We choose a 3-way covering of the parameters \textit{ego vehicle initial speed, ego vehicle lateral position, pedestrian speed, agent vehicle model} as they are intuitively the most critical parameters for the collision avoidance performance, and a 2-way covering of the other parameters.
These settings result in 195 covering array tests generated by the ACTS tool \cite{kuhn2013introduction}.
Depending on the time budget for the testing, the strength of the covering array parameters can be changed, which can drastically increase or decrease the number of discrete test cases in the generated covering array.
We first execute the resulting 195 covering array tests and collect simulation trajectories.
}
Then, we compute the \textit{robustness} values for those trajectories, with respect to the requirements $R1$, $R2$ and $R4$.
Finally, for each requirement, starting from the case with the smallest positive robustness value, we try to find as many additional falsifications as possible, within a maximum of 300 extra simulations, by using a falsification approach that uses simulated annealing to perform the optimization.

For requirement $R1$, 67 cases were falsified from the covering array tests (i.e., 67 of the 195 cases did not satisfy $R1$).
Starting from 7 of the remaining (non-falsifying) cases from the covering array tests, 5 additional falsifying cases were discovered using falsification.
For requirement $R2$, 65 cases were falsified from the covering array cases, with an additional 8 cases discovered during the falsification step. For requirement $R4$, 67 cases were falsified during the covering array step, with 12 more cases discovered during the falsification step.

These results demonstrate that we can automatically identify test cases  that violate specific sensor-level, system-level, and sensor-to-system level requirements.
These test cases can be fed back to the designers to improve the perception or control design or can be used as guidance to identify challenging scenarios to be used during the testing phase. 


\subsubsection*{Analysis of robustness values on the falsification of Model $M_2$}
The \textit{robustness} value, which is described in Sec. \ref{sec:preliminaries}, for a trajectory with respect to the requirement is automatically computed in Sim-ATAV.
This computation is performed by the S-TaLiRo tool \cite{AnnapureddyLFS11tacas} and is used to guide the test cases towards a falsification.

\begin{figure}[]
	\begin{centering}
		\includegraphics[trim={0.25in 3.5in 6in 0.1in}, clip,width=\columnwidth]{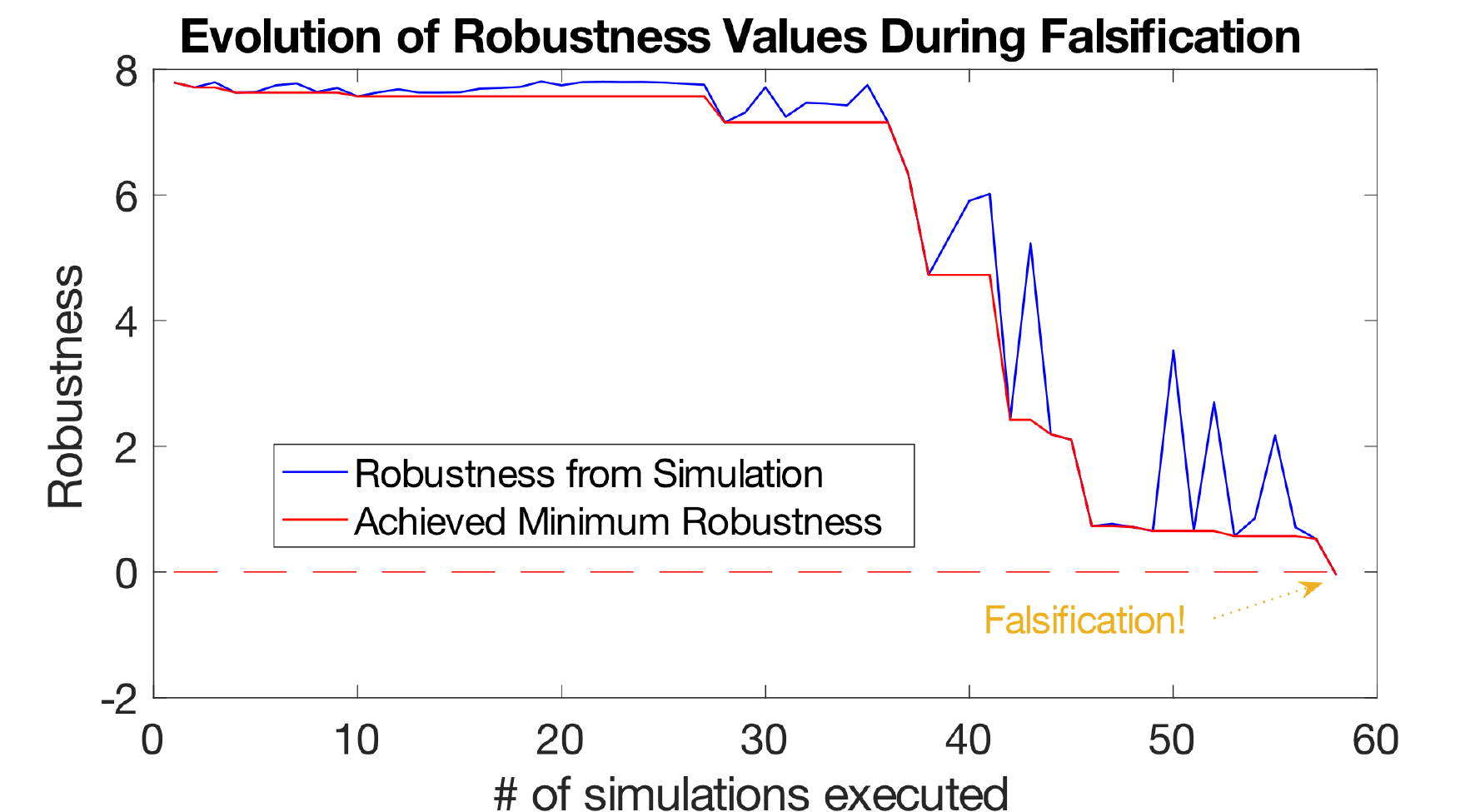}
		\caption{Robustness guided falsification utilizes global optimization techniques to guide the test cases toward falsification.}
		\label{fig:scenario_2_r4_rob_evolution}
	\end{centering}
\end{figure}

We use the results of falsification on Model $M_2$ to show, in Fig. \ref{fig:scenario_2_r4_rob_evolution}, how the robustness value changes over time and finally becomes negative, which indicates falsification of the requirement.
In this case, Sim-ATAV was able to find a falsifying example in $58$ simulations.
Because the cost function gradients are not computable, we use a stochastic global optimization technique, Simulated Annealing (SA).
The blue line shows the robustness value for each simulation. We can observe that the robustness value per simulation run is not monotonic. This is due to the stochastic nature of the optimizer;
however, the achieved minimum robustness up to the current simulation is a non-increasing function, which shows the best robustness achieved after each simulation.
As soon as the framework finds a test case that causes a negative robustness value, it stops the search and reports the falsifying example.

\begin{figure}[b]
	\begin{center}
		\vspace*{-0.125in}
		\hspace*{-0.125in}
		\begin{tabular}{c c}
			\includegraphics[width=.48\columnwidth]{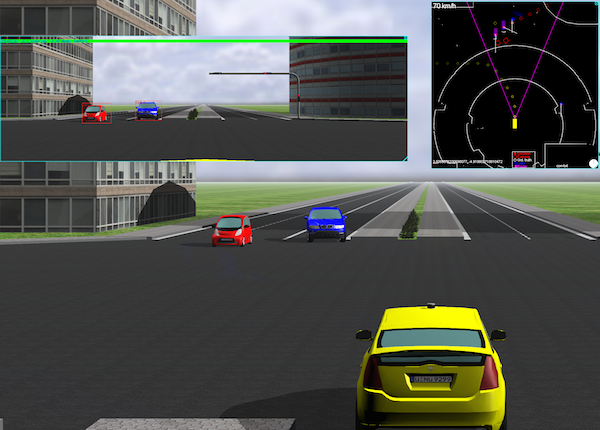}\hspace*{0in}
			&\includegraphics[width=.48\columnwidth]{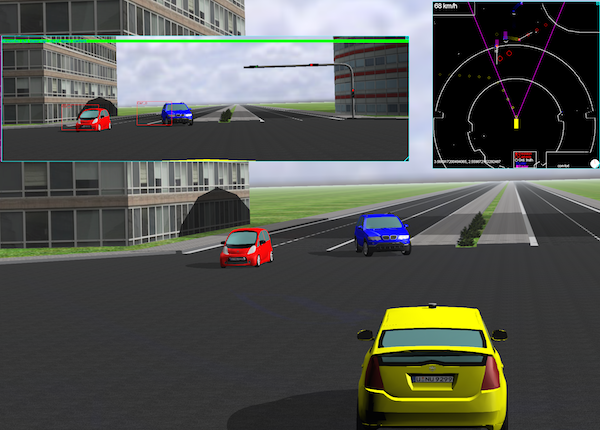}\hspace*{0in}\\
			(a) Perception error & (b) Perception error\\
			\includegraphics[width=.48\columnwidth]{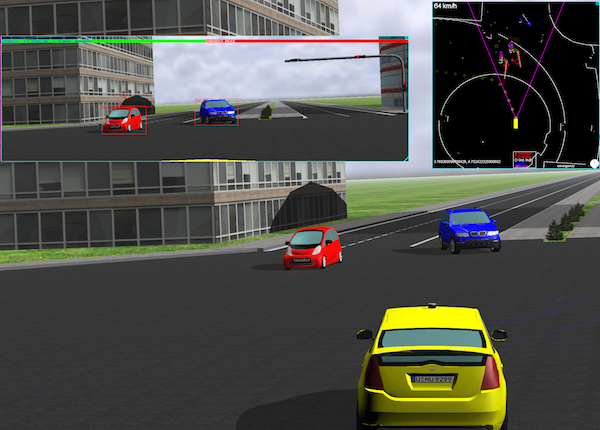}\hspace*{0in}
			&\includegraphics[width=.48\columnwidth]{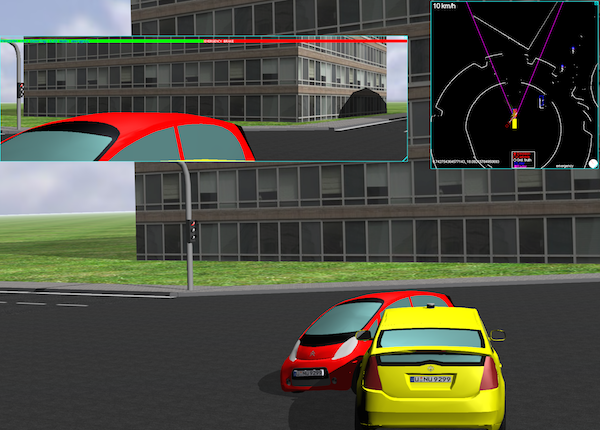}\hspace*{0in}\\
			(c) Emergency braking & (d) Collision
		\end{tabular}
		\caption{Time-ordered images from the falsifying example on model $M_1$.}
		\label{fig:scenario2_images}
	\end{center}
\end{figure}

Fig. \ref{fig:scenario2_images} shows images from the simulation execution of a falsifying example for model $M_2$ with respect to the requirement $R2$.
Between the time corresponding to Fig. \ref{fig:scenario2_images}-(a) to Fig. \ref{fig:scenario2_images}-(b), the red car approaching from the opposite side is driving on a path such that there will be a future collision with the Ego vehicle.
However, due to incorrect localization of the agent vehicle, the Ego vehicle is not able to correctly predict the future trajectory of the agent vehicle, and so it does not predict a collision.
Hence, it continues without taking action to avoid the collision.
Starting from the moment shown on Fig. \ref{fig:scenario2_images}-(c), the Ego vehicle predicts the collision and starts applying emergency braking;
however, because it takes action too late, the Ego vehicle cannot avoid a collision with the agent vehicle, as shown in Fig. \ref{fig:scenario2_images}-(d).

We note that, even for cases that are non-falsifying, the robustness values are useful for the system designer's analysis, as behaviors with low robustness value are ``close to" violating the requirement and, therefore, correspond to cases that may require closer attention.

\subsubsection*{A visual analysis of a falsifying simulation trajectory from Model $M_3$}
As presented in Table~\ref{table:experiment_results}, Sim-ATAV was able to find a falsifying example for model $M_3$ with respect to the STL requirement \perfreq\ in $232$ simulations.
We present a visual analysis of the falsifying test result.
Note that this analysis is done automatically in the framework, and corresponding satisfaction/falsification of the requirement is returned to the user, along with the \textit{robustness} value that shows the signed distance to the boundary of satisfaction or falsification.
The type of visual analysis we present here may be useful for the system designers to understand the reason behind the falsification (or satisfaction) of a requirement, which can be helpful for debugging or improving the design.
For this analysis, we use the definitions and notation introduced in Section \ref{sec:stl_specs_for_experiments}.

\begin{figure}[]
	\begin{centering}
		\includegraphics[trim={0.25in 0.1in 1.1in 0.25in}, clip,width=\columnwidth]{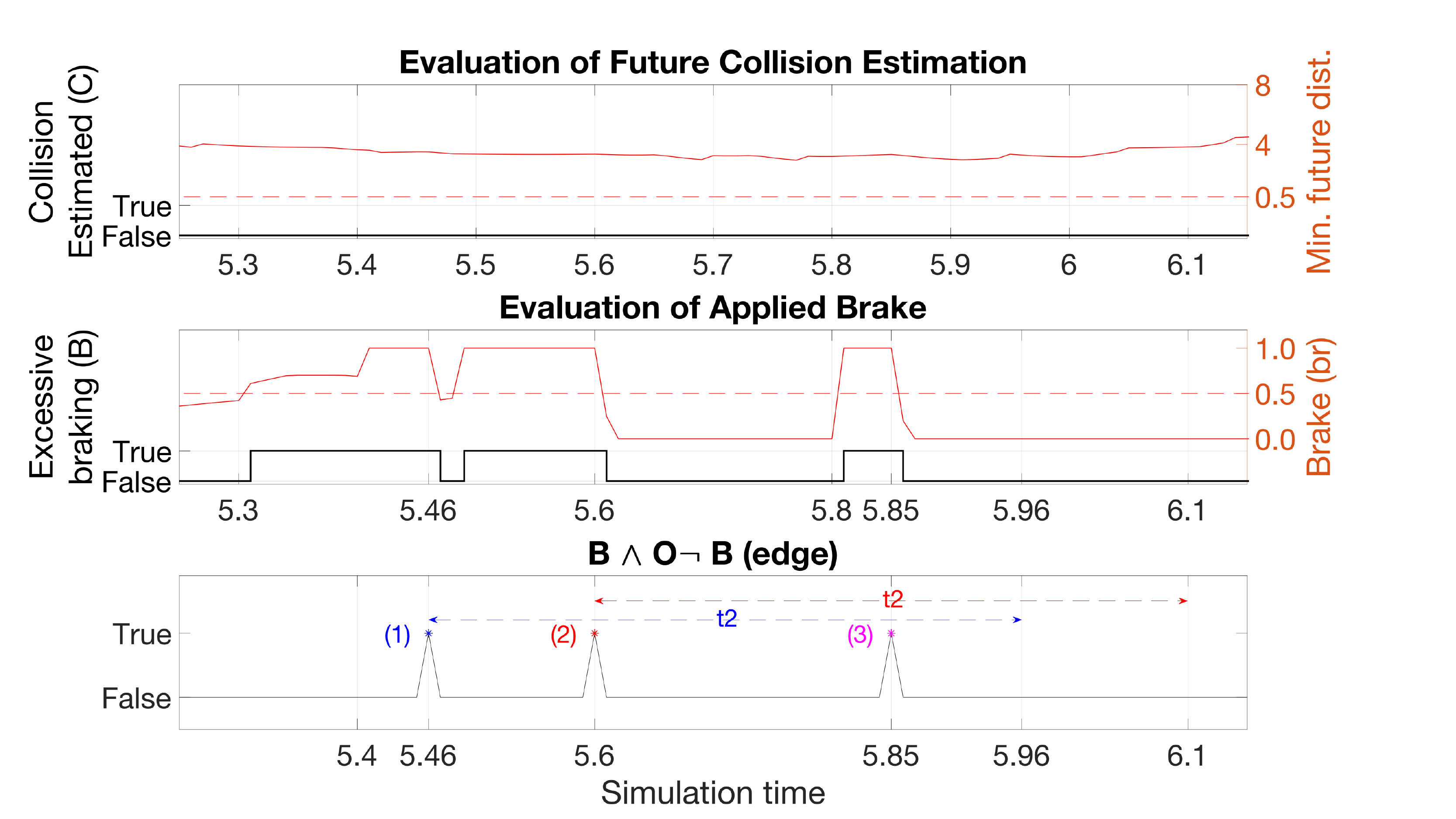}
		\caption{Analysis of falsification for Model $M_3$.}
		\label{fig:scenario3analysis}
	\end{centering}
\end{figure}

Fig. \ref{fig:scenario3analysis} shows a part of the simulation trajectory of Model $M_3$ for a time window around the falsification instance, together with the corresponding logic evaluations of the predicates related to the subformulas in Requirement \perfreq.
In the top plot in Fig. \ref{fig:scenario3analysis}, the red solid line is the estimated future minimum distance between the ego vehicle and Agent vehicle 1, with respect to the simulation time.
This estimation is based on the ground truth information collected from the simulation and utilizes the CTRV model at each time step of the simulation to compute the collision estimate that is described in \perfreq.
For this example, we define the variable $\collestimated$ that is used in \perfreq\  as $(d_{f, min} < 0.5)$, where $d_{f, min}$ represents the expected minimum future distance.
The dashed horizontal red line in Fig. \ref{fig:scenario3analysis} located at $0.5m$ is the threshold minimum future distance for a collision prediction.
The values of $t1$ and $t2$ are respectively defined as $0.6$ and $0.5$ in this example.
Since $d_{f, min}$ is never less than $0.5$ in this case, the collision estimation variable $\collestimated$, which is represented by the black solid line in the top plot, is always false.

The middle plot presents a similar evaluation for computing the variable $B$ used in \perfreq, which represents excessive braking.
This evaluation uses the collected actual normalized brake power data, say $br$, from the simulation and computes the logical variable $B = (br > 0.5)$.
The solid and dashed red lines represent $br$ and the threshold value $0.5$, respectively.
The solid black line shows the value of $B$ with respect to time.
The bottom plot in Fig. \ref{fig:scenario3analysis} shows the value of the variable \edge\ that is defined for the requirement \perfreq\ with respect to the simulation time.

The first part of the requirement \perfreq, which was defined as $\Box \Big(\neg \Box_{[0,t1]}(B\wedge\neg \collestimated)\Big)$ in Section \ref{sec:stl_specs_for_experiments}, would evaluate to false if and only if there exists a time window of $t1$ such that $B$ is always true and $\collestimated$ is always false.
Focusing on the values of $\collestimated$ and $B$ from the top two plots in Fig. \ref{fig:scenario3analysis}, we can see that although $\collestimated$ is always false, because there is no time window of $t1=0.6s$ in which $B$ is always true, the first part of the requirement evaluates to true.
This means this execution of model $M_3$ satisfies the first part of the requirement \perfreq.

The second part of the requirement \perfreq, which is defined as $\Box \Big(\neg\big( \edge \wedge \Diamond_{(0,t2]}( \edge \wedge \Diamond_{(0,t2]}\edge ) \big) \Big)$ evaluates to false if and only if 
there exists a series of three brake releases (\edge), such that one occurrence of \edge\ follows another within a $t2$ time window.
As we see in the bottom plot of Fig. \ref{fig:scenario3analysis}, at time $5.6s$ it is true that there exists \edge\ and it is also true that there exists another \edge\ within the time window of $0$ to $0.5s$ following this moment (occurring at $5.85s$).
Hence, the inner $( \edge \wedge \Diamond_{(0,t2]}\edge )$ inside the above formula evaluates to true at time $5.6s$.
If we call this event $e1$, the overall formula will evaluate to false if there exists an \edge\ that is followed by event $e1$ in a time window between $0$ and $t2=0.5s$.
This happens at time $5.46s$, which is the moment that there exists an \edge\ followed by event $e1$ at $0.14 \in (0, 0.5]$, \textit{where the event $e1$ is defined as an \edge\ followed by another \edge\ within $t \in (0, 0.5]$}. 
Hence, the second part of the requirement \perfreq\ evaluates to false, and as a result, \perfreq\ evaluates to false at time $5.46s$, since it is a conjunction of parts 1 and 2.
In other words, the system falsifies (does not satisfy) the requirement \perfreq.

\section{Conclusions} \label{sec:conclusions}

We demonstrated a simulation-based adversarial test generation framework for autonomous vehicles.
The framework works in a closed-loop fashion, where the system evolves in time with the feedback cycles from the autonomous vehicle's controller.
The framework includes models of lidar and radar sensor behaviors, as well as a model of the CCD camera sensor inputs.
CCD camera images are rendered synthetically by our framework and processed using a pre-trained deep neural network (DNN). Using our framework, we demonstrated a new effective way of finding a critical vehicle behaviors by using 1) covering arrays to test combinations of discrete parameters and 2) simulated annealing to find corner-cases.

Future work will include using identified counterexamples to retrain and improve the DNN-based perception system, e.g., \cite{DreossiGYKSS2018ijcai}.
Additionally, the scene rendering will be made more realistic by using other scene rendering tools, such as those based on state-of-the-art game engines, e.g., \cite{Dosovitskiy2017crl}.
Also, we note that the formal requirements that we considered 
were provided as an example of the type used
when employing a requirements-driven development approach based on a temporal logic language, which is a formalism that may be unfamiliar to many 
test engineers.  
Future research will investigate ways to elicit formal requirements based for ADS using visual specification languages, e.g., \cite{HoxhaMF15iros}.





\bibliographystyle{IEEEtran}

\begin{thebibliography}{10}
\providecommand{\url}[1]{#1}
\csname url@rmstyle\endcsname
\providecommand{\newblock}{\relax}
\providecommand{\bibinfo}[2]{#2}
\providecommand\BIBentrySTDinterwordspacing{\spaceskip=0pt\relax}
\providecommand\BIBentryALTinterwordstretchfactor{4}
\providecommand\BIBentryALTinterwordspacing{\spaceskip=\fontdimen2\font plus
\BIBentryALTinterwordstretchfactor\fontdimen3\font minus
  \fontdimen4\font\relax}
\providecommand\BIBforeignlanguage[2]{{%
\expandafter\ifx\csname l@#1\endcsname\relax
\typeout{** WARNING: IEEEtran.bst: No hyphenation pattern has been}%
\typeout{** loaded for the language `#1'. Using the pattern for}%
\typeout{** the default language instead.}%
\else
\language=\csname l@#1\endcsname
\fi
#2}}

\bibitem{NHTSA2016}
``{NHTSA Federal Automated Vehicles Policy},''
  \url{https://www.transportation.gov/AV/federal-automated-vehicles-policy-september-2016},
  accessed: 2018-09-11.

\bibitem{SAE2015}
{SAE}, ``Guidelines for safe on-road testing of sae level 3, 4, and 5 prototype
  automated driving systems ({ADS}), document j3018\_201503,''
  \url{https://www.sae.org/standards/content/j3018\_201503/}, accessed:
  2018-09-12.

\bibitem{Christensen2015}
A.~Christensen, A.~Cunningham, J.~Engelman, C.~Green, C.~Kawashima, S.~Kiger,
  D.~Prokhorov, L.~Tellis, B.~Wendling, and F.~Barickman, ``Key considerations
  in the development of driving automation systems,'' in \emph{24th Enhanced
  Safety of Vehicles Conferences}, 2015.

\bibitem{NCAP2018}
``{Euro NCAP},'' \url{https://www.euroncap.com/en}, accessed: 2018-09-11.

\bibitem{Pegasus2018}
``{Pegasus},'' \url{https://www.pegasusprojekt.de/en/home}, accessed:
  2018-09-11.

\bibitem{Stellet2015TestingOA}
J.~E. Stellet, M.~R. Zofka, J.~Schumacher, T.~Schamm, F.~Niewels, and J.~M.
  Z{\"o}llner, ``Testing of advanced driver assistance towards automated
  driving: A survey and taxonomy on existing approaches and open questions,''
  \emph{2015 IEEE 18th International Conference on Intelligent Transportation
  Systems}, pp. 1455--1462, 2015.

\bibitem{Zofka2018}
M.~R. Zofka, M.~Essinger, T.~Fleck, R.~Kohlhaas, and J.~M. Z{\"{o}}llner, ``The
  sleepwalker framework: Verification and validation of autonomous vehicles by
  mixed reality lidar stimulation,'' in \emph{{SIMPAR}}.\hskip 1em plus 0.5em
  minus 0.4em\relax {IEEE}, 2018, pp. 151--157.

\bibitem{Koopman2016}
\BIBentryALTinterwordspacing
P.~Koopman and M.~Wagner, ``Challenges in autonomous vehicle testing and
  validation,'' \emph{SAE Int. J. Trans. Safety}, vol.~4, pp. 15--24, 04 2016.
  [Online]. Available: \url{https://doi.org/10.4271/2016-01-0128}
\BIBentrySTDinterwordspacing

\bibitem{Burgard2016}
\BIBentryALTinterwordspacing
W.~Burgard, U.~Franke, M.~Enzweiler, and M.~Trivedi, ``{The Mobile Revolution -
  Machine Intelligence for Autonomous Vehicles (Dagstuhl Seminar 15462)},''
  \emph{Dagstuhl Reports}, vol.~5, no.~11, pp. 62--70, 2016. [Online].
  Available: \url{http://drops.dagstuhl.de/opus/volltexte/2016/5764}
\BIBentrySTDinterwordspacing

\bibitem{Wachenfeld2016}
W.~Wachenfeld, P.~Junietz, R.~Wenzel, and H.~Winner, ``The
  worst-time-to-collision metric for situation identification,'' in \emph{2016
  {IEEE} Intelligent Vehicles Symposium, {IV} 2016, Gotenburg, Sweden, June
  19-22, 2016}, 2016, pp. 729--734.

\bibitem{geiger2012we}
A.~Geiger, P.~Lenz, and R.~Urtasun, ``Are we ready for autonomous driving? the
  kitti vision benchmark suite,'' in \emph{Computer Vision and Pattern
  Recognition (CVPR), 2012 IEEE Conference on}.\hskip 1em plus 0.5em minus
  0.4em\relax IEEE, 2012.

\bibitem{wu2017squeezedet}
B.~Wu, F.~Iandola, P.~H. Jin, and K.~Keutzer, ``{SqueezeDet}: Unified, small,
  low power fully convolutional neural networks for real-time object detection
  for autonomous driving,'' in \emph{IEEE Conference on Computer Vision and
  Pattern Recognition Workshops {(CVPRW)}}, 2017, pp. 446--454.

\bibitem{pomerleau1989alvinn}
D.~A. Pomerleau, ``Alvinn: An autonomous land vehicle in a neural network,'' in
  \emph{Advances in neural information processing systems}, 1989, pp. 305--313.

\bibitem{chen2015deepdriving}
C.~Chen, A.~Seff, A.~Kornhauser, and J.~Xiao, ``{DeepDriving}: Learning
  affordance for direct perception in autonomous driving,'' in
  \emph{Proceedings of the IEEE International Conference on Computer Vision},
  2015, pp. 2722--2730.

\bibitem{Chi2017}
L.~Chi and Y.~Mu, ``Deep steering: Learning end-to-end driving model from
  spatial and temporal visual cues,'' \emph{arXiv preprint arXiv: 1708.03798},
  2017.

\bibitem{StricklandFBA2018icra}
M.~Strickland, G.~Fainekos, and H.~B. Amor, ``Deep predictive models for
  collision risk assessment in autonomous driving,'' in \emph{IEEE
  International Conference on Robotics and Automation (ICRA)}, 2018.

\bibitem{krizhevsky2012imagenet}
A.~Krizhevsky, I.~Sutskever, and G.~E. Hinton, ``Imagenet classification with
  deep convolutional neural networks,'' in \emph{Advances in neural information
  processing systems}, 2012, pp. 1097--1105.

\bibitem{TianEtAl2018icse}
Y.~Tian, K.~Pei, S.~Jana, and B.~Ray, ``Deeptest: Automated testing of
  deep-neural-network-driven autonomous cars,'' in \emph{40th International
  Conference on Software Engineering ({ICSE})}, 2018.

\bibitem{PapernotEtAl2017asiaccs}
N.~Papernot, P.~McDaniel, I.~Goodfellow, S.~Jha, Z.~B. Celik, and A.~Swami,
  ``Practical black-box attacks against machine learning,'' in \emph{ACM Asia
  Conference on Computer and Communications Security}.\hskip 1em plus 0.5em
  minus 0.4em\relax ACM, 2017.

\bibitem{WickerHK2018tacas}
M.~Wicker, X.~Huang, and M.~Kwiatkowska, ``Feature-guided black-box safety
  testing of deep neural networks,'' in \emph{International Conference on Tools
  and Algorithms for the Construction and Analysis of Systems ({TACAS})}, ser.
  LNCS, vol. 10805, 2018, pp. 408--426.

\bibitem{hartman2005software}
A.~Hartman, ``Software and hardware testing using combinatorial covering
  suites,'' \emph{Graph theory, combinatorics and algorithms}, vol.~34, pp.
  237--266, 2005.

\bibitem{AbbasFSIG13tecs}
H.~Abbas, G.~E. Fainekos, S.~Sankaranarayanan, F.~Ivancic, and A.~Gupta,
  ``Probabilistic temporal logic falsification of cyber-physical systems,''
  \emph{ACM Transactions on Embedded Computing Systems}, vol.~12, no.~s2, May
  2013.

\bibitem{Tuncali2018}
C.~E. Tuncali, G.~Fainekos, H.~Ito, and J.~Kapinski, ``{Simulation-based
  Adversarial Test Generation for Autonomous Vehicles with Machine Learning
  Components},'' in \emph{{ {IEEE} Intelligent Vehicles Symposium ({IV})}},
  2018.

\bibitem{AnnapureddyLFS11tacas}
Y.~S.~R. Annapureddy, C.~Liu, G.~E. Fainekos, and S.~Sankaranarayanan,
  ``{S-TaLiRo}: {A} tool for temporal logic falsification for hybrid systems,''
  in \emph{Tools and algorithms for the construction and analysis of systems},
  ser. LNCS, vol. 6605.\hskip 1em plus 0.5em minus 0.4em\relax Springer, 2011,
  pp. 254--257.

\bibitem{NajmEtAl2013dot}
W.~G. Najm, R.~Ranganathan, G.~Srinivasan, J.~S. Toma, E.~Swanson, and A.~B.~D.
  Smith, ``Description of light-vehicle pre-crash scenarios for safety
  applications based on vehicle-to-vehicle communications,'' DOT HS 811 731,
  Tech. Rep., 2013.

\bibitem{ZhaoGJ17itsc}
D.~Zhao, Y.~Guo, and Y.~J. Jia, ``{TrafficNet}: An open naturalistic driving
  scenario library,'' in \emph{20th {IEEE} International Conference on
  Intelligent Transportation Systems, {ITSC}}, 2017.

\bibitem{ZhaoEtAl2017its}
D.~Zhao, H.~Lam, H.~Peng, S.~Bao, D.~J. LeBlanc, K.~Nobukawa, and C.~S. Pan,
  ``Accelerated evaluation of automated vehicles safety in lane-change
  scenarios based on importance sampling techniques,'' \emph{IEEE Transactions
  on Intelligent Transportation Systems}, vol.~18, no.~3, pp. 595--607, 2017.

\bibitem{LoosPN11fm}
S.~M. Loos, A.~Platzer, and L.~Nistor, ``Adaptive cruise control: Hybrid,
  distributed, and now formally verified,'' in \emph{Formal Methods}, ser.
  LNCS, vol. 6664.\hskip 1em plus 0.5em minus 0.4em\relax Springer, 2011, pp.
  42--56.

\bibitem{AlthoffEtAl2010ivs}
M.~Althoff, D.~Althoff, D.~Wollherr, and M.~Buss, ``Safety verification of
  autonomous vehicles for coordinated evasive maneuvers,'' in \emph{IEEE
  Intelligent Vehicles Symposium}, 2010.

\bibitem{TuncaliPF16itsc}
C.~E. Tuncali, T.~P. Pavlic, and G.~Fainekos, ``Utilizing {S-TaLiRo} as an
  automatic test generation framework for autonomous vehicles,'' in \emph{IEEE
  Intelligent Transportation Systems Conference}, 2016.

\bibitem{TuncaliF19itsc}
C.~E. Tuncali and G.~Fainekos, ``Rapidly-exploring random trees for testing
  automated vehicles,'' in \emph{{IEEE} Intelligent Transportation Systems
  Conference ({ITSC})}, 2019.

\bibitem{AlthoffL2018iv}
M.~Althoff and S.~Lutz, ``Automatic generation of safety-critical test
  scenarios for collision avoidance of road vehicles,'' in \emph{IEEE
  Intelligent Vehicles Symposium {(IV)}}, 2018.

\bibitem{TuncaliEtAl2017case}
C.~E. Tuncali, S.~Yaghoubi, T.~P. Pavlic, and G.~Fainekos, ``Functional
  gradient descent optimization for automatic test case generation for vehicle
  controllers,'' in \emph{IEEE International Conference on Automation Science
  and Engineering}, 2017.

\bibitem{OKelly2017}
M.~O'Kelly, H.~Abbas, and R.~Mangharam, ``Computer-aided design for safe
  autonomous vehicles,'' in \emph{2017 Resilience Week (RWS)}, 2017.

\bibitem{KimKDS17esl}
B.~Kim, Y.~Kashiba, S.~Dai, and S.~Shiraishi, ``Testing autonomous vehicle
  software in the virtual prototyping environment,'' \emph{Embedded Systems
  Letters}, vol.~9, no.~1, pp. 5--8, 2017.

\bibitem{KimJSSY16emsoft}
B.~Kim, A.~Jarandikar, J.~Shum, S.~Shiraishi, and M.~Yamaura, ``The {SMT}-based
  automatic road network generation in vehicle simulation environment,'' in
  \emph{International Conference on Embedded Software {(EMSOFT)}}.\hskip 1em
  plus 0.5em minus 0.4em\relax {ACM}, 2016, pp. 18:1--18:10.

\bibitem{FanQM2018ieeedt}
C.~Fan, B.~Qi, and S.~Mitra, ``Data-driven formal reasoning and their
  applications in safety analysis of vehicle autonomy features,'' \emph{IEEE
  Design Test}, vol.~35, no.~3, pp. 31--38, June 2018.

\bibitem{KellyAM2016}
M.~O'Kelly, H.~Abbas, S.~Gao, S.~Shiraishi, S.~Kato, and R.~Mangharam,
  ``{APEX}: Autonomous vehicle plan verification and execution,'' in
  \emph{{SAE} World Congress}, 2016.

\bibitem{ciresan11}
D.~Cire\c{s}an, U.~Meier, J.~Masci, and J.~Schmidhuber, ``A committee of neural
  networks for traffic sign classification,'' in \emph{Proceedings of the 2011
  International Joint Conference on Neural Networks (IJCNN)}, 2011, pp.
  1918--1921.

\bibitem{john14}
V.~John, K.~Yoneda, B.~Qi, Z.~Liu, and S.~Mita, ``Traffic light recognition in
  varying illumination using deep learning and saliency map,'' in
  \emph{Proceedings of the 2014 IEEE 17th International Conference on
  Intelligent Transportation Systems (ITSC)}, 2014.

\bibitem{angelova15icra}
A.~Angelova, A.~Krizhevsky, and V.~Vanhoucke, ``Pedestrian detection with a
  large-field-of-view deep network,'' in \emph{Proceedings of the IEEE
  International Conference on Robotics and Automation (ICRA '15)}, 2015.

\bibitem{DuttaEtAl2018adhs}
S.~Dutta, S.~Jha, S.~Sankaranarayanan, and A.~Tiwari, ``Learning and
  verification of feedback control systems using feedforward neural networks,''
  in \emph{Analysis and Design of Hybrid Systems}, 2018.

\bibitem{DreossiDS2017nfm}
T.~Dreossi, A.~Donze, and S.~A. Seshia, ``Compositional falsication of
  cyber-physical systems with machine learning components,'' in \emph{{NASA}
  Formal Methods ({NFM})}, ser. LNCS, vol. 10227.\hskip 1em plus 0.5em minus
  0.4em\relax Springer, 2017, pp. 357--372.

\bibitem{Dreossi2018}
T.~Dreossi, S.~Jha, and S.~A. Seshia, ``Semantic adversarial deep learning,''
  vol. 10981, pp. 3--26, 2018.

\bibitem{Abbas17Cyphy}
H.~Abbas, M.~O'Kelly, A.~Rodionova, and R.~Mangharam, ``Safe at any speed: A
  simulation-based test harness for autonomous vehicles,'' in \emph{$7^{th}$
  International Workshop on Cyber-Physical Systems {(CyPhy)}}, 2017.

\bibitem{Alur15book}
R.~Alur, \emph{Principles of Cyber-Physical Systems}.\hskip 1em plus 0.5em
  minus 0.4em\relax MIT Press, 2015.

\bibitem{AbbasHFU14cyber}
H.~Abbas, B.~Hoxha, G.~Fainekos, and K.~Ueda, ``Robustness-guided temporal
  logic testing and verification for stochastic cyber-physical systems,'' in
  \emph{IEEE International Conference on CYBER Technology in Automation,
  Control, and Intelligent Systems}, 2014.

\bibitem{BartocciEtAl2018survey}
E.~Bartocci, J.~Deshmukh, A.~Donz\'{e}, G.~Fainekos, O.~Maler, D.~Nickovic, and
  S.~Sankaranarayanan, ``Specification-based monitoring of cyber-physical
  systems: A survey on theory, tools and applications,'' in \emph{Lectures on
  Runtime Verification - Introductory and Advanced Topics}, ser. LNCS.\hskip
  1em plus 0.5em minus 0.4em\relax Springer, 2018, vol. 10457, pp. 128--168.

\bibitem{FainekosP09tcs}
G.~E. Fainekos and G.~J. Pappas, ``Robustness of temporal logic specifications
  for continuous-time signals,'' \emph{Theoretical Computer Science}, vol. 410,
  no.~42, pp. 4262--4291, 2009.

\bibitem{BoydV_book04}
S.~Boyd and L.~Vandenberghe, \emph{Convex Optimization}.\hskip 1em plus 0.5em
  minus 0.4em\relax Cambridge University Press, 2004.

\bibitem{HenzingerKPV98jcss}
T.~A. Henzinger, P.~W. Kopke, A.~Puri, and P.~Varaiya, ``What's decidable about
  hybrid automata?'' \emph{J. Comput. Syst. Sci.}, vol.~57, no.~1, pp. 94--124,
  1998.

\bibitem{HoxhaEtAl14difts}
B.~Hoxha, H.~Bach, H.~Abbas, A.~Dokhanchi, Y.~Kobayashi, and G.~Fainekos,
  ``Towards formal specification visualization for testing and monitoring of
  cyber-physical systems,'' in \emph{International Workshop on Design and
  Implementation of Formal Tools and Systems}, 2014.

\bibitem{KapinskiEtAl2016csm}
J.~Kapinski, J.~V. Deshmukh, X.~Jin, H.~Ito, and K.~Butts, ``Simulation-based
  approaches for verification of embedded control systems: An overview of
  traditional and advanced modeling, testing, and verification techniques,''
  \emph{IEEE Control Systems Magazine}, vol.~36, no.~6, pp. 45--64, 2016.

\bibitem{kuhn2013introduction}
D.~R. Kuhn, R.~N. Kacker, and Y.~Lei, \emph{Introduction to combinatorial
  testing}.\hskip 1em plus 0.5em minus 0.4em\relax CRC press, 2013.

\bibitem{ShalevSS2017arxiv}
S.~Shalev-Shwartz, S.~Shammah, and A.~Shashua, ``On a formal model of safe and
  scalable self-driving cars,'' arXiv:1708.06374v2, Tech. Rep., 2017.

\bibitem{Schubert2008}
R.~Schubert, E.~Richter, and G.~Wanielik, ``Comparison and evaluation of
  advanced motion models for vehicle tracking,'' in \emph{2008 11th
  International Conference on Information Fusion}, June 2008, pp. 1--6.

\bibitem{LeBlanc2013}
D.~J. LeBlanc, M.~Gilbert, S.~Stachowski, D.~Blower, C.~A.~C. Flannagan,
  S.~Karamihas, and W.~T.~B. andRini Sherony, ``Advanced surrogate target
  development for evaluating pre-collision systems,'' in \emph{23rd Enhanced
  Safety of Vehicles Conferences}, 2013.

\bibitem{Petrovskaya2009}
A.~Petrovskaya and S.~Thrun, ``Model based vehicle detection and tracking for
  autonomous urban driving,'' \emph{Autonomous Robots}, vol.~26, no.~2, pp.
  123--139, Apr 2009.

\bibitem{Grabe2009}
B.~Grabe, T.~Ike, and M.~H{\"{o}}tter, ``Evidence based evaluation method for
  grid-based environmental representation,'' in \emph{{FUSION}}.\hskip 1em plus
  0.5em minus 0.4em\relax {IEEE}, 2009, pp. 1234--1240.

\bibitem{ISO2018}
``{ISO Functional Safety},'' \url{https://en.wikipedia.org/wiki/ISO\_26262},
  accessed: 2018-09-11.

\bibitem{Elbanhawi2015}
M.~Elbanhawi, M.~Simic, and R.~Jazar, ``In the passenger seat: Investigating
  ride comfort measures in autonomous cars,'' \emph{IEEE Intelligent
  Transportation Systems Magazine}, vol.~7, no.~3, pp. 4--17, Fall 2015.

\bibitem{Green2016}
P.~Green, ``Motion sickness and concerns for self-driving vehicles: A
  literature review,''
  \url{http://umich.edu/~driving/publications/Motion-Sickness--Report-061616pg-sent.pdf},
  accessed: 2018-09-11.

\bibitem{dokhanchi2016efficient}
A.~Dokhanchi, B.~Hoxha, C.~E. Tuncali, and G.~Fainekos, ``An efficient
  algorithm for monitoring practical tptl specifications,'' in \emph{Formal
  Methods and Models for System Design (MEMOCODE), 2016 ACM/IEEE International
  Conference on}.\hskip 1em plus 0.5em minus 0.4em\relax IEEE, 2016, pp.
  184--193.

\bibitem{michel2004cyberbotics}
O.~Michel, ``Cyberbotics ltd. {Webots}: professional mobile robot simulation,''
  \emph{International Journal of Advanced Robotic Systems}, vol.~1, no.~1,
  p.~5, 2004.

\bibitem{abadi2016tensorflow}
M.~Abadi, A.~Agarwal, P.~Barham, E.~Brevdo, Z.~Chen, C.~Citro, G.~S. Corrado,
  A.~Davis, J.~Dean, M.~Devin, \emph{et~al.}, ``Tensorflow: Large-scale machine
  learning on heterogeneous distributed systems,'' \emph{arXiv
  preprint:1603.04467}, 2016.

\bibitem{ester1996density}
M.~Ester, H.-P. Kriegel, J.~Sander, X.~Xu, \emph{et~al.}, ``A density-based
  algorithm for discovering clusters in large spatial databases with noise.''
  in \emph{Kdd}, vol.~96, no.~34, 1996, pp. 226--231.

\bibitem{schubert2017dbscan}
E.~Schubert, J.~Sander, M.~Ester, H.~P. Kriegel, and X.~Xu, ``Dbscan revisited,
  revisited: why and how you should (still) use dbscan,'' \emph{ACM
  Transactions on Database Systems (TODS)}, vol.~42, no.~3, p.~19, 2017.

\bibitem{wan2000unscented}
E.~A. Wan and R.~Van Der~Merwe, ``The unscented kalman filter for nonlinear
  estimation,'' in \emph{Proceedings of the IEEE 2000 Adaptive Systems for
  Signal Processing, Communications, and Control Symposium (Cat. No.
  00EX373)}.\hskip 1em plus 0.5em minus 0.4em\relax Ieee, 2000, pp. 153--158.

\bibitem{donze10cav}
A.~Donze, ``Breach, a toolbox for verification and parameter synthesis of
  hybrid systems,'' in \emph{Computer Aided Verification {(CAV)}}, ser. LNCS,
  vol. 6174.\hskip 1em plus 0.5em minus 0.4em\relax Springer, 2010, pp.
  167--170.

\bibitem{DreossiGYKSS2018ijcai}
T.~Dreossi, S.~Ghosh, X.~Yue, K.~Keutzer, A.~L. Sangiovanni{-}Vincentelli, and
  S.~A. Seshia, ``Counterexample-guided data augmentation,'' in
  \emph{Proceedings of the Twenty-Seventh International Joint Conference on
  Artificial Intelligence, {IJCAI}}, 2018, pp. 2071--2078.

\bibitem{Dosovitskiy2017crl}
A.~Dosovitskiy, G.~Ros, F.~Codevilla, A.~Lopez, and V.~Koltun, ``{CARLA}: {An}
  open urban driving simulator,'' in \emph{Proceedings of the 1st Annual
  Conference on Robot Learning}, 2017, pp. 1--16.

\bibitem{HoxhaMF15iros}
B.~Hoxha, N.~Mavridis, and G.~Fainekos, ``{VISPEC}: {A} graphical tool for
  elicitation of {MTL} requirements,'' in \emph{Proceedings of the IEEE/RSJ
  International Conference on Intelligent Robots and Systems}, 2015.

\end{thebibliography}

\ifthenelse {\boolean{inclAuthors}}
{
%

\vspace*{-0.3in}

\begin{IEEEbiography}[{\includegraphics[width=1in,height=1.25in,clip,keepaspectratio]{./authorphotos/erkan.jpg}}]{Cumhur Erkan Tuncali}
(S'15) received his B.S. degree in Electronics Engineering from Hacettepe University, Ankara, Turkey, in 2004 and his M.Sc. degree in Electrical and Electronics engineering from Middle East Technical University, Ankara, Turkey, in 2007.
He received a Ph.D. degree in Computer Engineering from Arizona State University, Tempe, AZ, USA in 2019.

After receiving his B.S. degree, he developed embedded systems and software for electronic payment and access control systems.
In 2009, he joined Turkish Aerospace Industries where he worked as an Avionics Software Design Specialist for a full-scale avionics modernization project.
During his Ph.D. studies, he worked as a Graduate Research Assistant in the Cyber-Physical Systems Lab. with Prof. Fainekos in Arizona State University.
As part of his Ph.D. work, he has also contributed to the research projects on autonomous driving at Toyota Motor North America as an intern R\&D engineer.

His main research work has been on the automatic test generation for cyber-physical systems with a focus on autonomous vehicles. 
His research interests also include robotics, neural networks and control systems. 

\end{IEEEbiography}

\begin{IEEEbiography}[{\includegraphics[width=1in,height=1.25in,clip,keepaspectratio]{./authorphotos/Geo.jpg}}]{Georgios Fainekos} 
(S'04 - M'08) received a Diploma degree (B.Sc. \& M.Sc.) in Mechanical Engineering from the National Technical University of Athens in 2001. He received an M.Sc. and a Ph.D. in Computer and Information Science from the University of Pennsylvania in 2004 and 2008. 

He is an Associate Professor at the School of Computing, Informatics and Decision Systems Engineering (SCIDSE) at Arizona State University (ASU). He is director of the Cyber-Physical Systems (CPS) Lab and he is currently affiliated with the NSF I/UCRC Center for Embedded Systems (CES) and the Robotics Faculty Group at ASU. 
His technical expertise is on applied logic, formal verification, testing, control theory, artificial intelligence, and optimization. His research has applications to automotive systems, medical devices, autonomous (ground and aerial) vehicles, and human-robot interaction (HRI).

In 2013, Dr. Fainekos received the NSF CAREER award and the ASU SCIDSE Best Researcher Junior Faculty Award. He is also recipient of the 2008 Frank Anger Memorial ACM SIGBED/SIGSOFT Student Award. 
\end{IEEEbiography}

\vspace*{-0.3in}

\begin{IEEEbiography}[{\includegraphics[width=1in,height=1.25in,clip,keepaspectratio]{./authorphotos/Danil.jpg}}]{Danil Prokhorov} (SM'02) started his research career
in Russia. He studied system engineering which
included courses in math, physics, mechatronics and
computer technologies, as well as aerospace and
robotics. He received his M.S. with Honors in St.
Petersburg, Russia, in 1992. After receiving Ph.D. in
1997, he joined the staff of Ford Scientific Research
Laboratory, Dearborn, Michigan. While at Ford he
pursued AI research focusing on neural networks
with applications to system modeling, powertrain
control, diagnostics and optimization. He has been
involved in research and planning for various intelligent technologies, such
as highly automated vehicles, AI and other futuristic systems at Toyota Tech
Center (TTC), Ann Arbor, MI since 2005. Since 2011 he is in charge of future
research department in Toyota Motor North America R \& D. 

Dr. Prokhorov has been
serving as a panel expert for NSF, DOE, ARPA, Associate Editor of several
scientific journals, and in the leadership of IEEE Intelligent Transportation
Systems and International Neural Network Society (INNS). He has authored
more than 100 scientific publications, as well as 70 patents. He has recently
been elected the INNS Fellow. Having shown feasibility of autonomous driving
and personal flying mobility, his department continues research of complex
multi-disciplinary problems while exploring opportunities for the next big
thing.
\end{IEEEbiography}

\vspace*{-0.5in}

\begin{IEEEbiography}[{\includegraphics[width=1in,height=1.25in,clip,keepaspectratio]{./authorphotos/Isaac.jpg}}]{Hisahiro Ito}
received a Ph.D. degree in computational engineering and science (polymer physics) from Nagoya University in 2003. 

He lead the development of the Software-in-the-Loop (SiL) simulator for Toyota's engine and transmission control software from 2009 to 2012. He is currently responsible for the integration of verification and validation techniques into development processes for large scale, complex control system models.
\end{IEEEbiography}

\begin{IEEEbiography}[{\includegraphics[width=1in,height=1.25in,clip,keepaspectratio]{./authorphotos/Jim.jpg}}]{James Kapinski} (SM'15)
received the B.S. and M.S. degrees in electrical engineering from the University of Pittsburgh in 1996 and 1999. He received the Ph.D. in electrical and computer engineering from Carnegie Mellon University (CMU) in 2005.

He was a postdoctoral researcher at CMU from 2007 to 2008. 
He went on to found and lead Fixed-Point Consulting, serving clients in the defense, aerospace, and automotive industries. Since 2012 he has been with Toyota Motor North America. He is currently serving as a Senior Principal Scientist in the Toyota Research Institute of North America in Ann Arbor, MI, USA. His work at Toyota focuses on advanced research into verification techniques for cyber-physical systems.
\end{IEEEbiography}







}
{}

\end{document}